\documentclass[letterpaper,twocolumn,fleqn]{article}
\usepackage{ist}
\usepackage[english]{babel}
\usepackage{amsmath}
\usepackage{amsfonts}
\usepackage{amssymb}
\usepackage{amsthm}
\usepackage{lscape}
\usepackage{multirow}
\usepackage{pdflscape}
\usepackage{relsize}
\usepackage{url}
\usepackage{cite}
\usepackage{graphicx}
\usepackage[titlenotnumbered,algoruled,longend,linesnumbered]{algorithm2e}
\usepackage{colortbl}

\newtheorem{definition}{Definition}
\newcommand{\placeimage}[2][0.95]{
  \includegraphics[draft=false,width=#1\textwidth,keepaspectratio]{#2}
}
\newcommand{\placeimagerotate}[2][0.95]{
  \includegraphics[draft=false,angle=90,width=#1\textwidth,keepaspectratio]{#2}
}
\title{
    Improved Image Selection for Stack-Based HDR Imaging
}
\author{
    Peter van Beek,
    Cheriton School of Computer Science,
    University of Waterloo, Canada
}
\date{}
\begin{document}
\maketitle
\thispagestyle{empty}

\begin{abstract}
Stack-based high dynamic range (HDR) imaging is a technique for
achieving a larger dynamic range in an image by combining
several low dynamic range images acquired at different
exposures. Minimizing the set of images to combine, while
ensuring that the resulting HDR image fully captures the scene's
irradiance, is important to avoid long image acquisition
and post-processing times. The problem of selecting the set
of images has received much attention. However, existing
methods either are not fully automatic, can be slow, or can
fail to fully capture more challenging scenes. In this paper,
we propose a fully automatic method for selecting the set of
exposures to acquire that is both fast and more accurate. We
show on an extensive set of benchmark scenes that our proposed
method leads to improved HDR images as measured against ground
truth using the mean squared error, a pixel-based metric,
and a visible difference predictor and a quality score, both
perception-based metrics.
\end{abstract}

%
%
%
%
%
\section{Introduction}

The sensor on a digital camera often cannot capture the
full dynamic range of a natural scene, resulting in either
dark, noisy regions or saturated regions in the image.
Stack-based high dynamic range (HDR) imaging is a technique
for achieving a larger dynamic range in an image by combining
several low dynamic range (LDR) images acquired at different
exposures~\cite{Mann1993,Debevec1997}. The resulting HDR
image can then be displayed on an HDR monitor, tonemapped for
display on a standard LDR monitor, used in computer vision
algorithms for object recognition, or used as a source of
illumination for computer-generated objects and scenes
\cite{Reinhard2010,Bloch2012}.

An essential part of a stack-based approach is selecting the
set of images to be combined. The set must be small, in order
to decrease capture and post-processing times, but must
accurately and fully capture the scene's irradiance.
The problem has received much attention in the
literature \cite{Chen2002,Grossberg2003,Barakat2008,Gelfand2010,Granados2010,Hasinoff2010,Hirakawa2010,Gallo2012,Sesh2012,Huang2013,Kehtarnavaz2015}.
A low-cost method to select the set of images that is used
in many cameras is to bracket around the best exposure
at a fixed progression. However, the method is not fully
automatic as the number of exposures $N$ to acquire must be
specified in advance, requiring information that is often not
available to the photographer. This shortcoming is shared
by several more elaborate methods that have been proposed
(e.g., \cite{Chen2002,Grossberg2003,Hirakawa2010}). As well,
restricting $N$ to be some small, fixed value (e.g., $N \le
3$, \cite{Gelfand2010,Huang2013,Kehtarnavaz2015}) will be
insufficient in more challenging HDR scenes, and setting $N$
to be a larger value would often acquire unnecessary images.

To accurately and fully capture a scene's irradiance,
ideally a method must be adaptive to the irradiance distribution
in a scene \cite{Granados2010,Gallo2012,Kehtarnavaz2015},
and incorporate
a model of camera noise \cite{Granados2010,Hasinoff2010}.
Barakat et al.~\cite{Barakat2008} and Hasinoff et
al.~\cite{Hasinoff2010} propose methods that use an estimate
of the extent of the dynamic range, and incorporate a noise
model that ensures each pixel is properly exposed in each
LDR image \cite{Barakat2008} or in the final HDR image
\cite{Hasinoff2010}.
Granados et al.~\cite{Granados2010},
Gallo et al.~\cite{Gallo2012}, and
Seshadrinathan et al.~\cite{Sesh2012}
propose methods that use
an estimate of the full irradiance distribution,
the HDR histogram, in their selection of images and
incorporate a noise model that ensures 
each pixel is properly exposed in
the final HDR image.
Pourreza-Shahri and Kehtarnavaz \cite{Kehtarnavaz2015} propose
a method for selecting images that is
adaptive to the irradiance distribution in a scene by clustering
the pixel values in a single well-exposed image.

As well, to be suitable in practice, a method must have low
computational complexity. Unfortunately, if we consider
just the state-of-the-art methods that are adaptive
to the irradiance distribution in the scene, only
Barakat et al.~\cite{Barakat2008} and
Pourreza-Shahri and Kehtarnavaz \cite{Kehtarnavaz2015}
propose methods whose
computational complexity is polynomial. The remaining
methods propose solving an optimization problem that is
either explicitly NP-complete (e.g., formulated as an integer
linear program \cite{Hasinoff2010}) or whose computational
complexity is unknown but empirically known to be high
\cite{Granados2010,Gallo2012,Sesh2012}.

In this paper, we propose a method for selecting the set
of exposures to acquire that is both fast and more accurate,
particularly on challenging scenes. Following previous work, our
proposed method is also adaptive to the irradiance distribution
and incorporates a model of camera noise. However, our method
formulates the selection problem as a \emph{polynomially}
solvable set covering problem. We show on a total of 110
benchmark scenes that overall, in addition to being fast,
our proposed method leads to improved HDR images over the
state-of-the-art methods as measured against ground truth
using the mean squared error, a pixel-based metric, and
a visible difference predictor and a quality score, both
perception-based metrics. Our experimental evaluation is also
the first to extensively evaluate existing state-of-the-art
methods for image selection for stack-based HDR imaging.

%
%
\section{Background}
\label{SECTION:Background}

In this section, we review some of the underlying concepts
and routines used in methods for selecting the set of images.

\begin{figure*}[t!]
\centering
\placeimage[1.00]{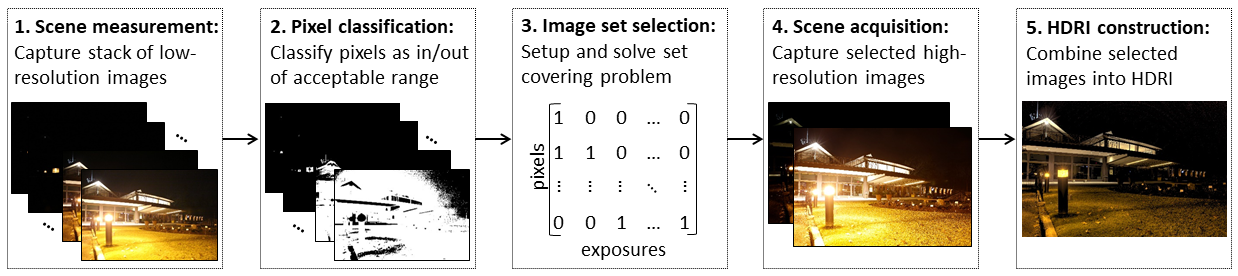}
\caption{
    Pipeline for proposed end-to-end system for image set
    selection for construction of HDR image of a scene.
}\label{FIGURE:pipeline}
\end{figure*}

%
%
Many cameras are able to record images in a proprietary RAW format in
addition to the common JPEG format. The RAW images are linearly
related to scene radiance \cite{Chakrabarti2009,Szeliski2010}
and have a higher dynamic range (usually, 12--14 bits), while
JPEG images are nonlinearly related to scene radiance and
have a lower dynamic range (8 bits).

\begin{figure}[b!]
\centering
\begin{tabular}{@{}c@{}c@{}}
\placeimage[0.24]{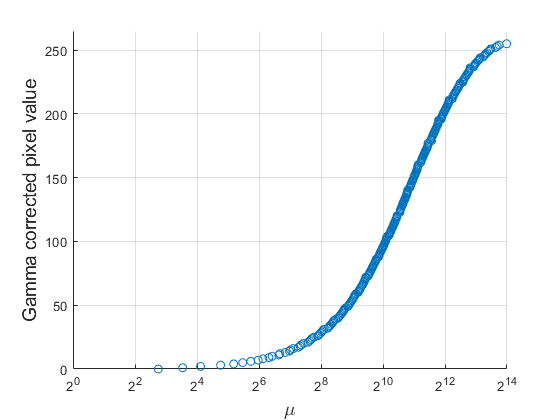} &
\placeimage[0.24]{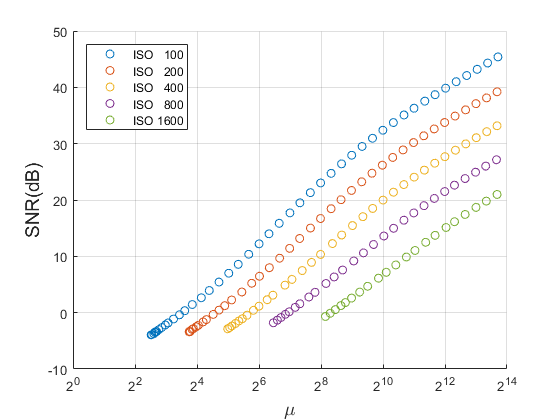} \\
(a) &
(b)
\end{tabular}
\caption{
    For a Canon EOS 5D Mark III camera,
    (a) radiometric response function,
    (b) signal to noise ratio expressed in decibels as a function of
    $\mu$, the RAW pixel value.
}\label{FIGURE:SNR}
\end{figure}

%
%
A camera radiometric response function $f$ is the nonlinear
mapping that determines how radiance in the scene becomes
pixel values in a JPEG image through the imaging pipeline
of the camera. In our experiments, a response function was
estimated using multiple pairs of RAW and JPEG images, all
taken of the same scene but each pair taken with a different
exposure \cite{Healey1994,Debevec1997,Mitsunaga1999,Kim2012}.
Alternatively, a response function can be roughly modeled using
a standard gamma correction curve. Once the mapping is known,
JPEG pixel values can be inverted back to an estimate of the
RAW pixel value using the inverse function $f^{-1}$. Figure~\ref{FIGURE:SNR}(a)
shows an example radiometric response function for a Canon
EOS 5D Mark III camera.

%
%
A camera noise level function $\mathrm{SNR(dB)}$ maps
a RAW pixel value to an estimate of the signal to
noise ratio \cite{Liu2008}. In our experiments, we followed Healey and
Kondepudy's~\cite{Healey1994} procedure for estimating camera
noise from multiple images, although methods have also been
proposed that estimate noise from a single image (see, e.g.,
\cite{Liu2008}). The signal to noise ratio expressed in
decibels is given by,
$\mathrm{SNR(dB)} = 20 \cdot \log_{10}( \mu/\sigma )$,
where $\mu$ is the RAW pixel value
and $\sigma$ is the estimated noise
at that RAW pixel value.
SNR increases monotonically (up to sensor saturation) with
increased exposure time and decreases monotonically with
increased ISO gain.
The SNR at saturation is zero by definition.
Figure~\ref{FIGURE:SNR}(b) shows example $\mathrm{SNR(dB)}$ curves
for a Canon EOS 5D Mark III camera.
The empirical data can be fit with negligible residual error
using the parametric noise model
$\sigma =
\sqrt{
    \mu g + r^2 g^2 + c^2
}$,
where
$g$ is the ISO gain,
$r$ is the read noise, and
$c$ is the noise component that does not depend on the signal
or the gain
(see, e.g., \cite{Healey1994,Hasinoff2010}).
For the Canon EOS 5D Mark III camera, $c$ is zero in the best fit.

%
%
The best methods for selecting the set of images
rely on either being given an estimate of the extent of the
dynamic range of a scene, or on being given an estimate of the
full irradiance distribution in the form of an HDR histogram.
In our experiments, we followed Gallo et al.'s~\cite{Gallo2012}
procedure for estimating the HDR histogram by first capturing
multiple low-resolution JPEG images from the live preview
stream by sweeping the lens through the available shutter
speeds and then combining the resulting LDR histograms into
a HDR histogram in the cumulative distribution function domain.

%
%
\section{Selection as Set Covering}

In this section, we present our method for selecting the set
of images. Our approach is inspired by the focus stacking
system proposed by Vaquero et al.~\cite{VaqueroGTPT2011}. For
simplicity of presentation and because of their advantages,
we assume that RAW images are to be selected and combined into
an HDR image, although our approach can also be adapted to
the selection of JPEG images. Our method proceeds as follows
(see Figure~\ref{FIGURE:pipeline}).

\emph{Step 1.}
Capture a stack of low-resolution JPEG images from the live
preview stream by sweeping the lens through the available
shutter speeds. Note that these images can be acquired for
``free'' if one is already estimating the extent of the
dynamic range or the HDR histogram (see last paragraph,
``Background'').

\emph{Step 2.}
Classify each pixel in each low resolution JPEG image as to
whether it is accurately captured. A pixel is accurately
captured by an exposure if its grayscale conversion falls
within the interval $[I_{\mathit{min}}, I_{\mathit{max}}]$,
where $I_{\mathit{min}}$ is the darkest pixel $p$ such that
$\mathrm{SNR(dB)}$ at $f^{-1}(p)$ is above a given threshold
and $I_{\mathit{max}}$ is the brightest pixel value such that
two or more component channels are rarely saturated.

\emph{Step 3.}
The key step is that we formulate the selection of images for
HDR imaging as a set covering problem.

\begin{definition}[Set covering]
Let $A = [a_{ij}]$ be an $m \times n$ (0-1)-matrix with a cost $w_j$
associated with each column. A row $i$
of $A$ is \emph{covered} by a column $j$ if
$a_{ij}$ is equal to one. The \emph{set covering
problem} is to find a subset of the columns $C \subseteq \{1,
\ldots, n\}$ that minimizes the total cost $\sum_{j \in C} w_j$
such that every row is covered; i.e., for every $i \in \{1,
\ldots, m\}$ there exists a $j \in C$ such that $a_{ij} = 1$.
\end{definition}

Let $\{t_1, \ldots, t_n\}$ be the ordered set of available
shutter speeds on a camera. In the set covering instance, a
row represents a pixel across the stack of low-resolution JPEG
images (Step 1), a column represents a possible exposure setting
$t_j$ for that pixel, and an entry $a_{ij}$ is 1 if and only if
the pixel has been accurately captured by that exposure (Step 2).
If the goal is to minimize the number of images selected,
$w_j = 1$, and if the goal is to minimize the total capture
time, $w_j = t_j + t_\mathit{over}$, where $t_\mathit{over}$
is the overhead between image acquisitions. In general,
solving set covering is NP-complete \cite{GareyJ79}. However,
here the set covering instance can be constructed to have the
consecutive ones property, allowing the selection of images to
be computed in polynomial time \cite{Nemhauser1988,Schobel2004}.

\begin{definition}[Consecutive ones property]
A set covering problem is said to have the \emph{consecutive
ones property} if the ones in each row of the matrix $A$
appear consecutively.
\end{definition}

If all costs $w_j$ are one, the following simple
reduction rules alone solve an instance.
Let $M_j = \{ i : a_{ij} = 1 \}$
and $N_i = \{ j : a_{ij} = 1 \}$, where 
$i \in \{1, \ldots, m\}$ and
$j \in \{1, \ldots, n\}$.

\begin{itemize}
\item[R1.]
If $N_{i_1} \subseteq N_{i_2}$, row $i_2$ can be removed.
\item[R2.]
If $M_{j_1} \subseteq M_{j_2}$ and $w_{j_1} \ge w_{j_2}$,
column $j_1$ can be removed.
\end{itemize}
If the costs are the shutter speeds, a polynomial
algorithm can be used to find a solution after reduction
\cite{Nemhauser1988,Schobel2004}.

\emph{Steps 4 \& 5.}
As the final steps in our proposed method
(see Figure~\ref{FIGURE:pipeline}),
acquire high-resolution RAW images at the exposures specified
by the solution to the set covering instance, and combine them
into a single HDR image using existing techniques
(e.g., \cite{Debevec1997,Kirk2006,Robertson2003}).

In the form stated above, our proposed method chooses an
exposure by setting the shutter speed, keeping the aperture
and ISO gain fixed. Keeping the aperture fixed at the camera's
native ISO is desirable in stack-based imaging to reduce image
noise and increase dynamic range (for the Canon EOS 5D Mark III
camera, the native ISO is 100; see Figure~\ref{FIGURE:SNR}(b)
for the effect of ISO on image noise and dynamic range).
However, increasing the ISO gain can be useful when the camera
is not mounted on a tripod and a minimum shutter speed may be
required to reduce the impact of camera shake, or to handle dynamic
scenes \cite{Gelfand2010,Hasinoff2010,Sesh2012}. Our proposed
set covering method can seamlessly handle a fixed but higher
ISO gain setting to ensure a minimum shutter speed: the low
resolution stack will be acquired at the higher ISO (Step 1),
the set covering will take into account the higher ISO when
setting the interval $[I_{\mathit{min}}, I_{\mathit{max}}]$
for determining whether a pixel has been accurately captured
(Step 2 \& 3), and the high resolution images will also be acquired
at the higher ISO (Step 4).

\section{Experimental Evaluation}

In this section, we experimentally evaluate the effectiveness
of our proposed method for image selection.

We compare our proposed set covering method against four
representative state-of-the-art methods:
(i) Barakat et al.~\cite{Barakat2008},
(ii) Hasinoff et al.~\cite{Hasinoff2010},
(iii) Pourreza-Shahri and Kehtarnavaz \cite{Kehtarnavaz2015}, and
(iv) Seshadrinathan et al.~\cite{Sesh2012}\footnote{The MATLAB
implementations and benchmark images are available at:
\url{https://anonymous}.}.
All five methods rely, to varying degrees, on knowing the
camera response function, the noise level function, and the
HDR histogram of the scene. The same routines were used across
all methods (see ``Background''). As well,
when needed a threshold of 20 dB was used for acceptable SNR,
and a range of $[20, 230]$ was used for determining whether a
pixel has been accurately captured. The grayscale pixel value
of 20 corresponds to an SNR of 20 dB for our Canon EOS 5D Mark
III camera and the value of 230 was empirically determined
to be the grayscale threshold where two or more component
channels are rarely saturated.
Pourreza-Shahri and Kehtarnavaz's \cite{Kehtarnavaz2015} method
requires the specification of a parameter $w$, where $w$ is used in clustering
the dark and bright regions of the well-exposed image. We
set $w = 8$, as in Pourreza-Shahri and Kehtarnavaz's \cite{Kehtarnavaz2015} experiments.
Similarly, Seshadrinathan et al.'s~\cite{Sesh2012} method requires
the specification of
$N$, where $N$ is an upper bound on how
many exposures to consider. Because of efficiency considerations,
Seshadrinathan et al.'s~\cite{Sesh2012} set $N$ to be three
in their experiments. In our experiments we set $N$ to be five;
larger values are impracticable.

We test the methods on the following benchmarks.
In each benchmark, only the shutter speed was varied
and the aperture and ISO gain were kept fixed.
\begin{itemize}
\item
The HDR Photographic Survey \cite{Fairchild2007}
suite consists of 105 benchmark image
sets\footnote{\url{http://www.rit-mcsl.org/fairchild/HDR.html}}.
Each image set consists of nine images (with one exception)
of a scene captured with a Nikon D2X camera. In each scene,
the exposure step was set to one stop and a $4288 \times 2848$
high resolution RAW image was acquired at each of the nine
shutter speeds using the camera's continuous auto-bracketing
function. We converted each RAW image to JPEG using Nikon's
ViewNX2 software and downsampled to give a $964 \times 640$
low resolution JPEG image that simulates the image that
would have been acquired from the live preview stream at that
shutter speed.
\item
We acquired five benchmark image sets using a camera remote
control application we implemented. A Canon EOS 5D Mark III
camera mounted on a tripod was tethered to a computer via a USB
cable and controlled by software that makes use of the Canon SDK
(Version 2.11). In each scene, the exposure step was set to
$1/3$ of a stop and a $5760 \times 3840$ high resolution RAW
image was acquired at each of the 55 possible shutter speeds. A
$960 \times 640$ low resolution JPEG image was also acquired
from the live preview stream at each shutter speed.
\end{itemize}

A ground truth HDR image was constructed using all of the
RAW high-resolution images in a benchmark (except if the
image would only add noise such as being fully saturated) and
compared against the HDR image constructed using only the RAW
images selected by each method. HDR images were constructed
using Photomatix Pro\footnote{\url{https://www.hdrsoft.com/}}.
We compare a method's HDR image, referred to as the \emph{test}
HDR image, against a \emph{ground truth} HDR image using the
following performance measures.
\begin{itemize}
\item
\emph{Quality correlate.}
The HDR-VDP-2.2.1 image quality metric, which quantifies the
visual distortion of the test HDR image from the ground truth
HDR image with a single quality score \cite{Mantiuk2011,Narwaria2015}.
\item
\emph{Visual difference prediction.}
The HDR-VDP-2.2.1 visual difference prediction metric, which
estimates the probability at each pixel that an average
human observer will detect a visually significant difference
between the test HDR image and the ground truth HDR image
\cite{Mantiuk2011,Narwaria2015}.
\item
\emph{Mean squared error.}
The mean squared error between the test HDR image and
the ground truth HDR image.
\end{itemize}

Our experimental evaluation is the first
to extensively evaluate existing state-of-the-art methods
for image selection for stack-based HDR imaging, with significantly
more methods, benchmarks, and performance measures being used compared
to the limited numbers used in previous evaluations.

\begin{figure}[t!]
\centering
\placeimage[0.41]{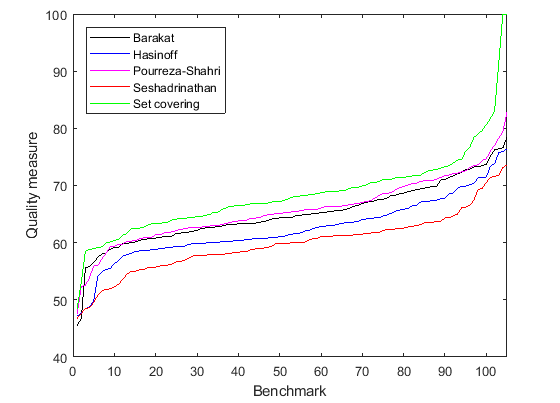} \\
(a) \\
\placeimage[0.41]{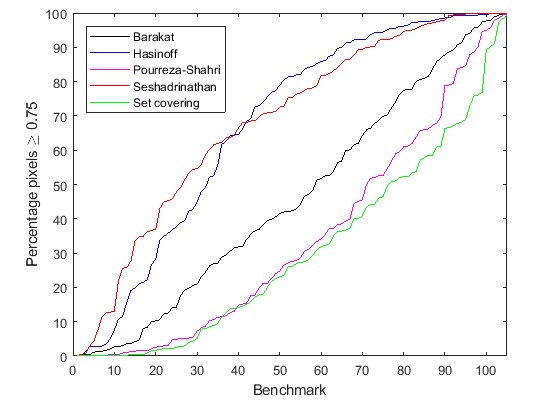} \\
(b) \\
\placeimage[0.41]{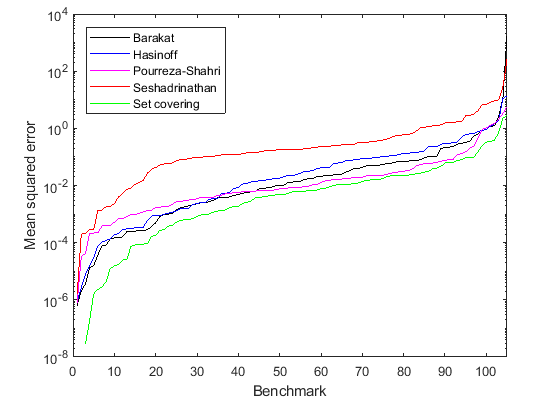} \\
(c)
\caption{
    For each method of selecting images and each benchmark in the
    HDR Photographic Survey:
    (a) quality correlate;
    (b) percentage of pixels with a probability $\ge 0.75$
	of a visually significant difference; and
    (c) mean squared error between test HDR images and ground truth
    HDR images. Within a method, a performance measure is sorted from smallest
    value to largest. Best viewed in color; our
    proposed set covering method has the highest quality and the lowest error.
}\label{FIGURE:PhotographicSurvey}
\end{figure}

Figure~\ref{FIGURE:PhotographicSurvey}(a) summarizes the \emph{quality}
of the test HDR images against the ground truth HDR images, as measured
by the quality correlate,
for the 105 HDR Photographic Survey benchmarks.
Figure~\ref{FIGURE:PhotographicSurvey}(b)\&(c) summarizes the
\emph{errors} for the 105 HDR Photographic Survey benchmarks. To
measure errors, for each benchmark we summarize the
probability at each pixel that an average human observer
will detect a visually significant difference from ground
truth with the percentage of pixels that are greater
than or equal to 0.75; i.e., the percentage of pixels
where a difference is very likely to be detected (see
Figure~\ref{FIGURE:PhotographicSurvey}(a)). As well, to measure
errors, for each benchmark we summarize the mean squared error
between the test HDR images and the ground truth HDR images
(see Figure~\ref{FIGURE:PhotographicSurvey}(b)). It can be
seen that on these benchmarks, even with the limited range of
choice in each benchmark---only nine images are available for
selection, each a full stop apart---our set covering method
achieves improvements across all three performance measures.

\begin{landscape}
\begin{figure}[thb]
\centering
\newcommand{\scale}{0.195}
\centering
\setlength{\tabcolsep}{4pt}
\begin{tabular}{@{}cccccc@{}}
\small
\placeimage[\scale]{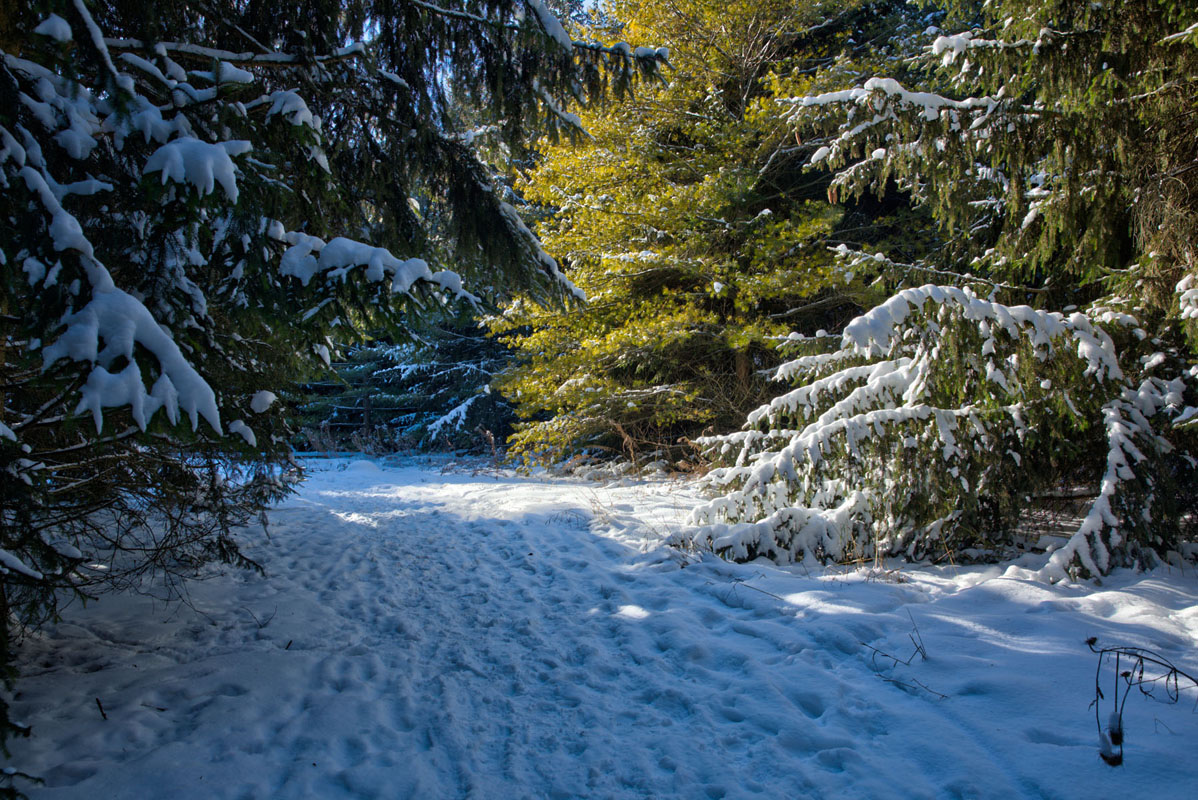}
    & \placeimage[\scale]{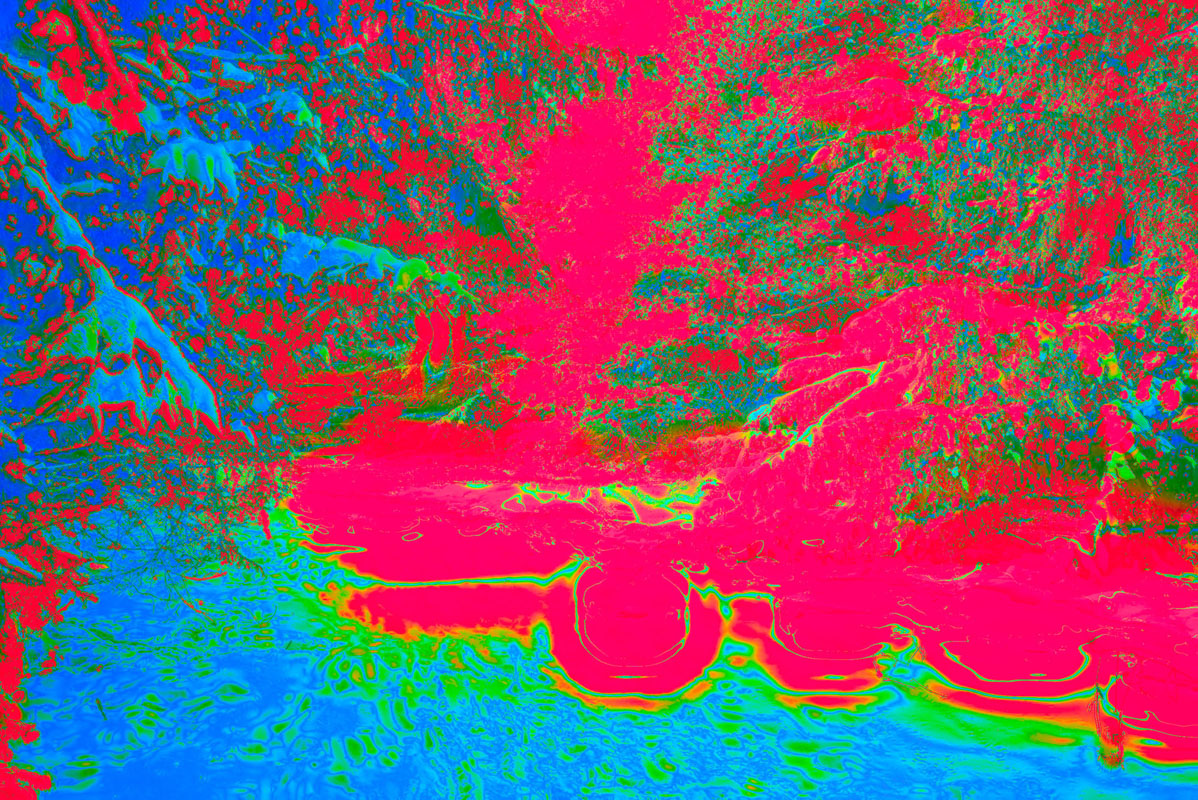}
    & \placeimage[\scale]{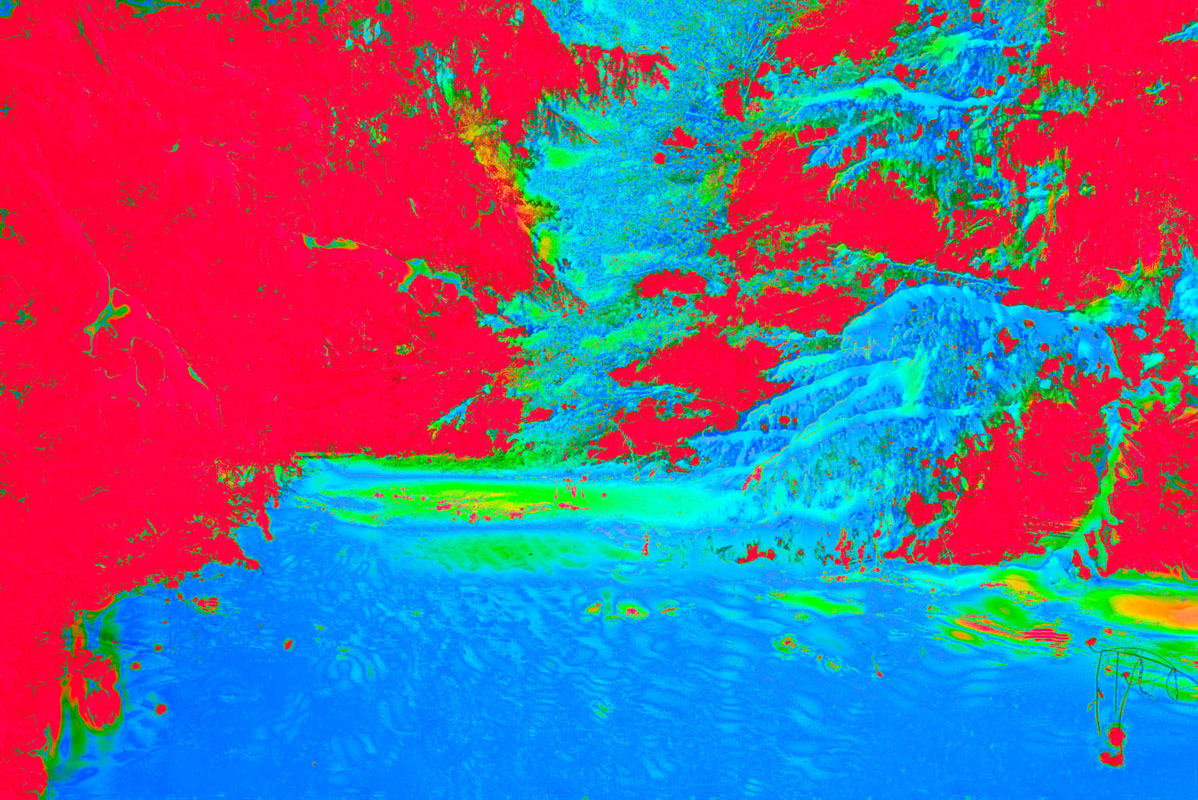}
    & \placeimage[\scale]{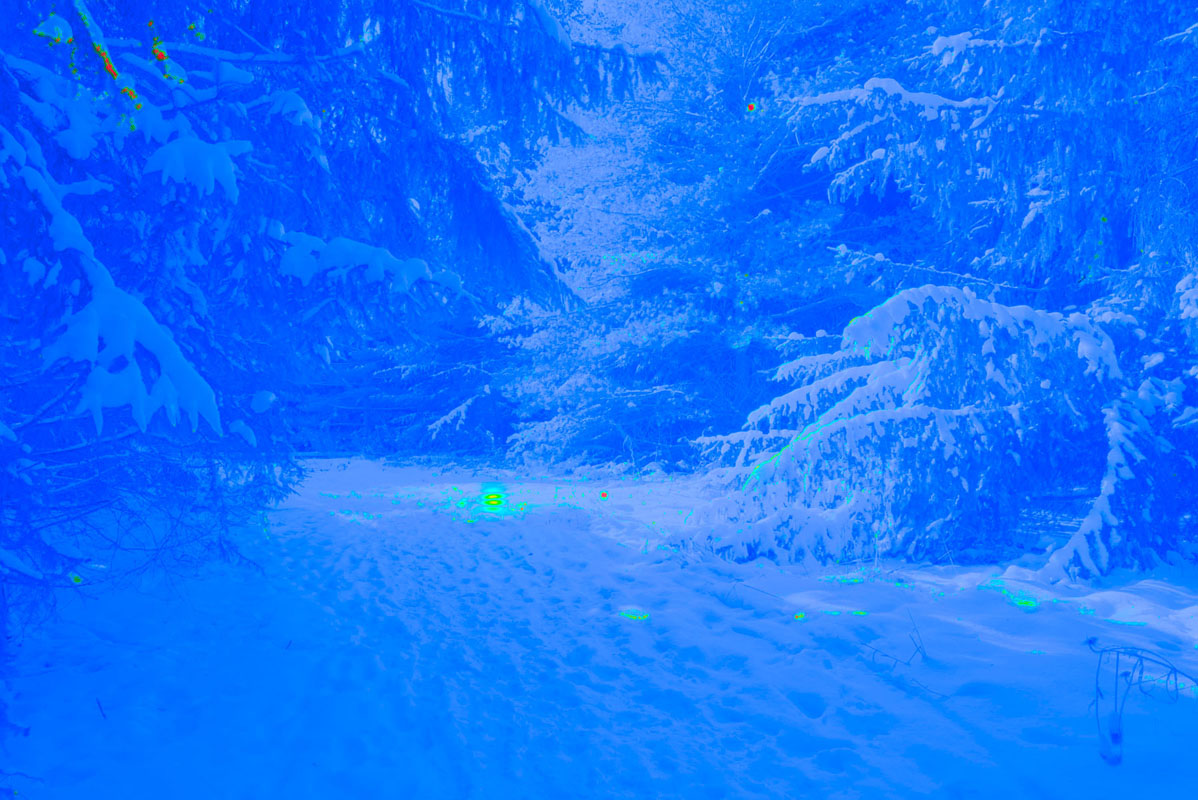}
    & \placeimage[\scale]{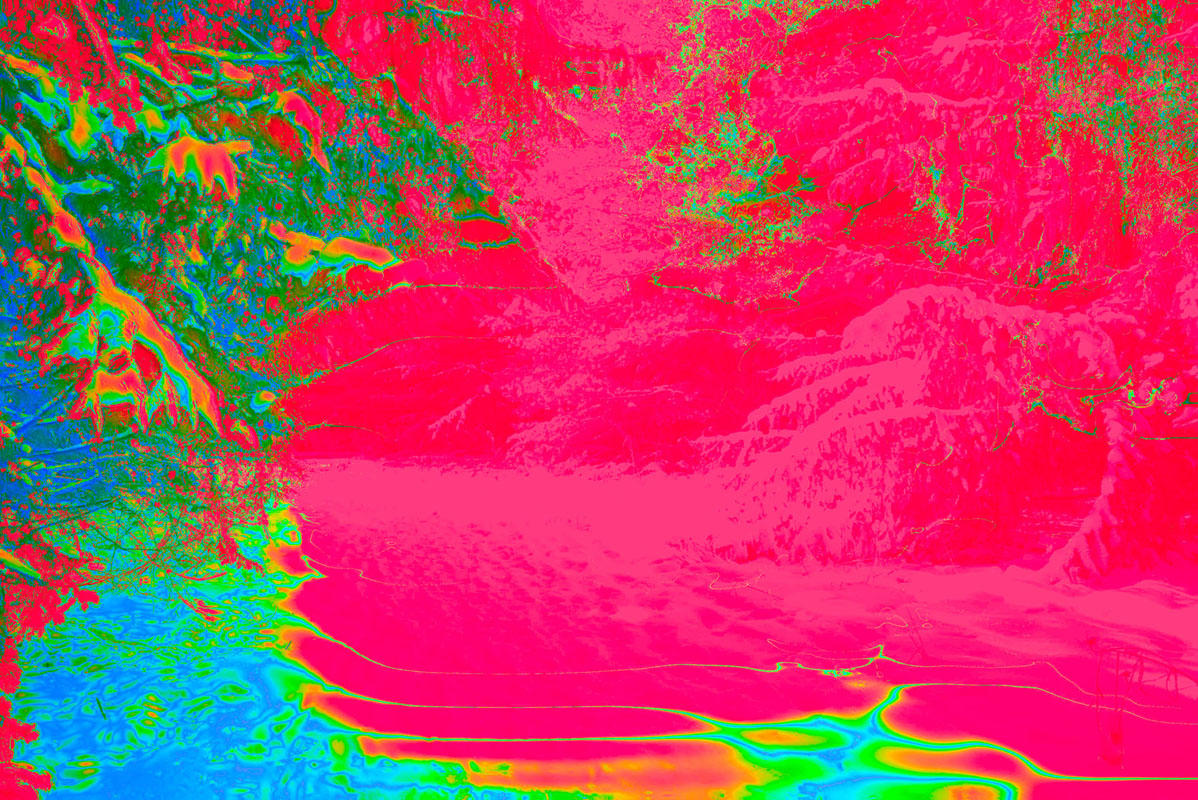}
    & \placeimage[\scale]{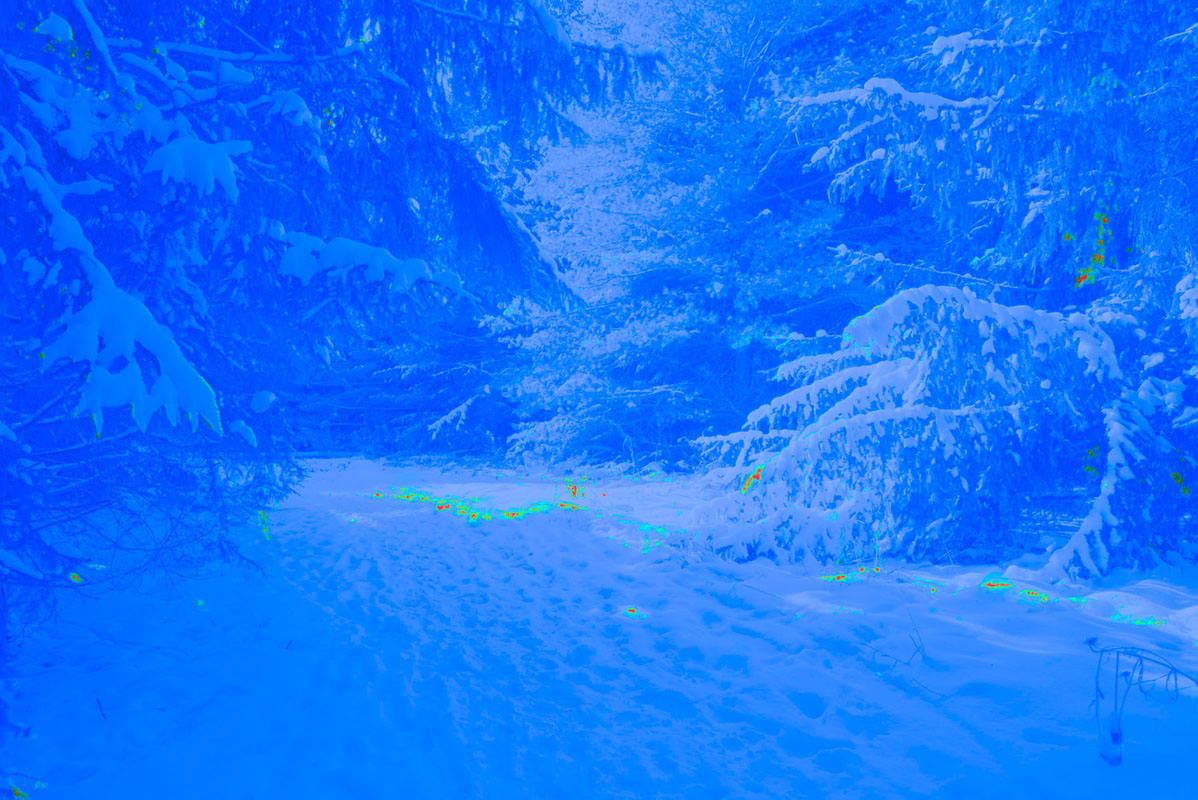} \\
    &  62.7, $ 0.658 \times 10^{-2}$, 3
    &  67.3, $ 0.248 \times 10^{-2}$, 3
    &  61.1, $ 0.726 \times 10^{-2}$, 3
    &  53.6, $13.4   \times 10^{-2}$, 1
    &  67.1, $ 0.105 \times 10^{-2}$, 4 \\
\placeimage[\scale]{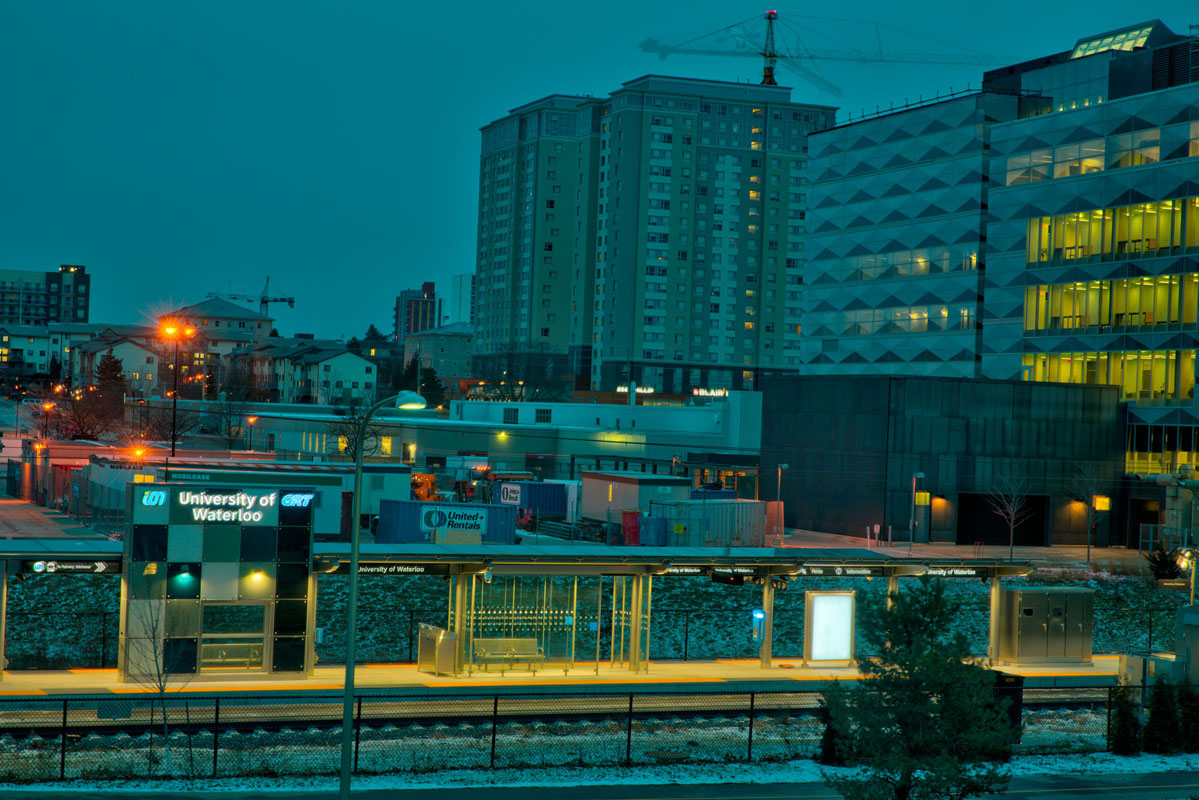}
    & \placeimage[\scale]{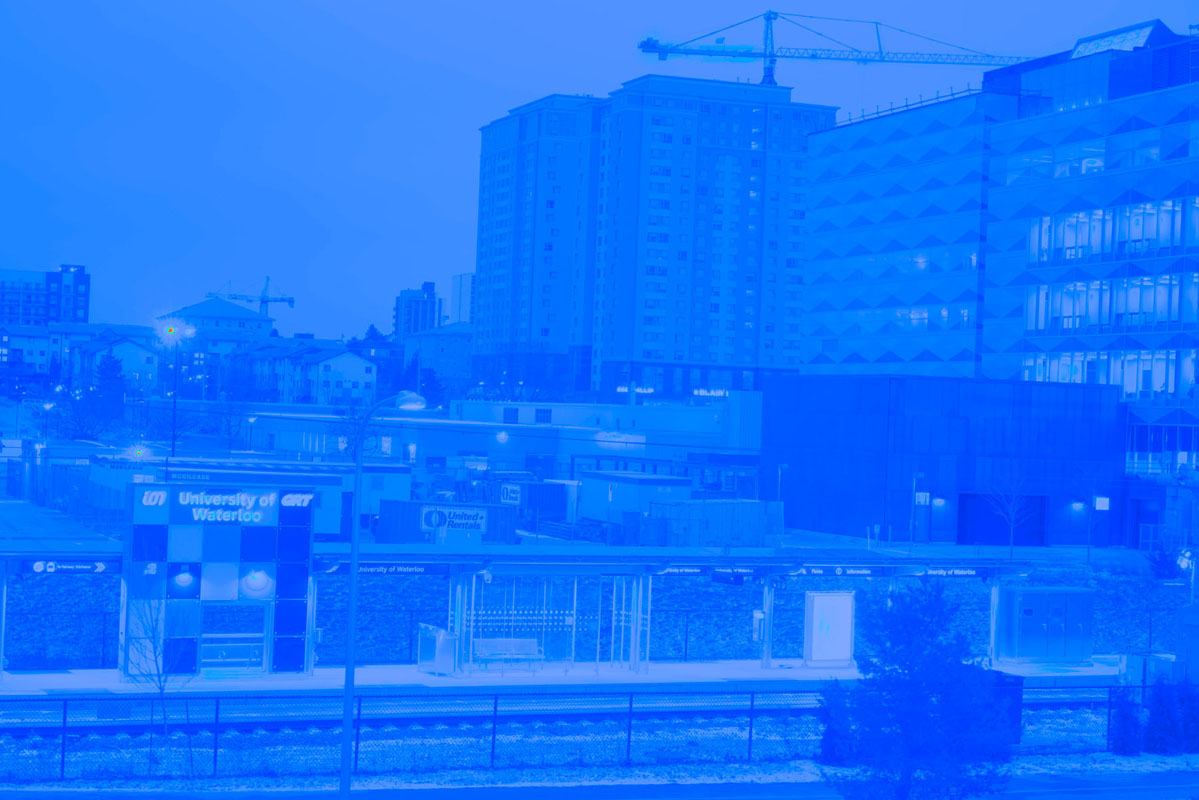}
    & \placeimage[\scale]{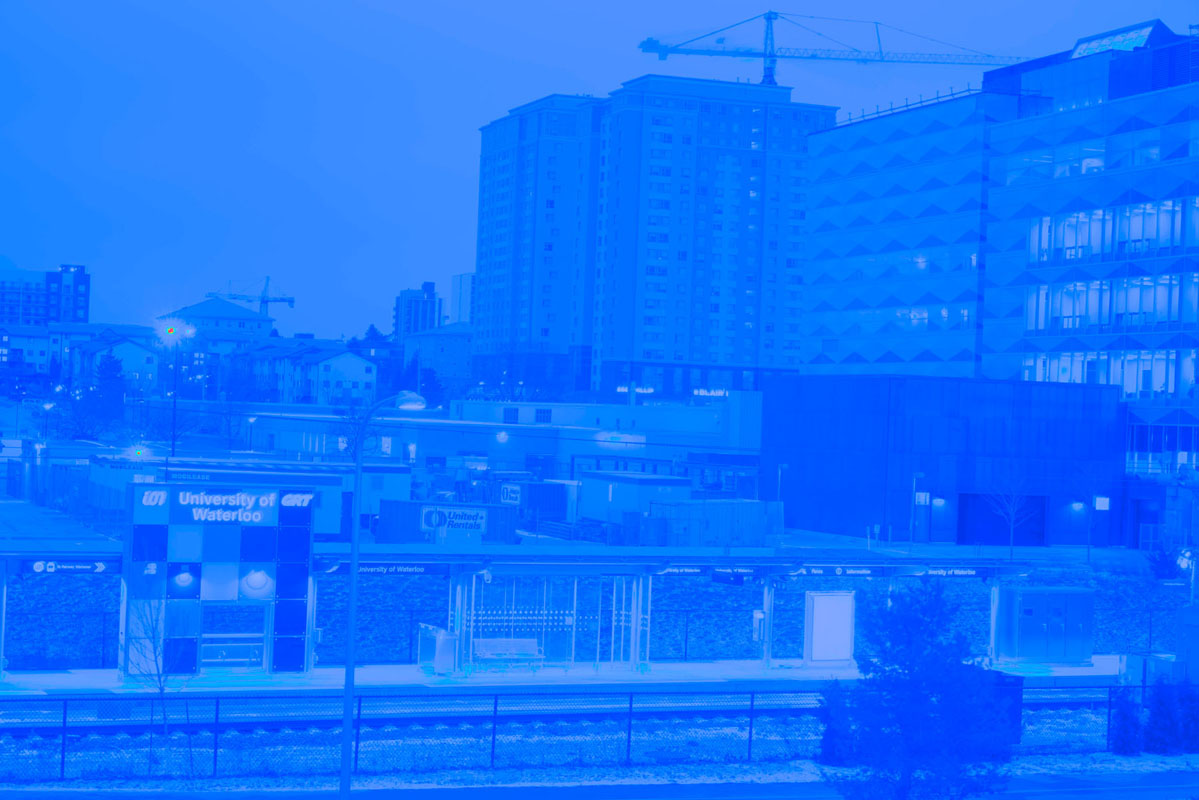}
    & \placeimage[\scale]{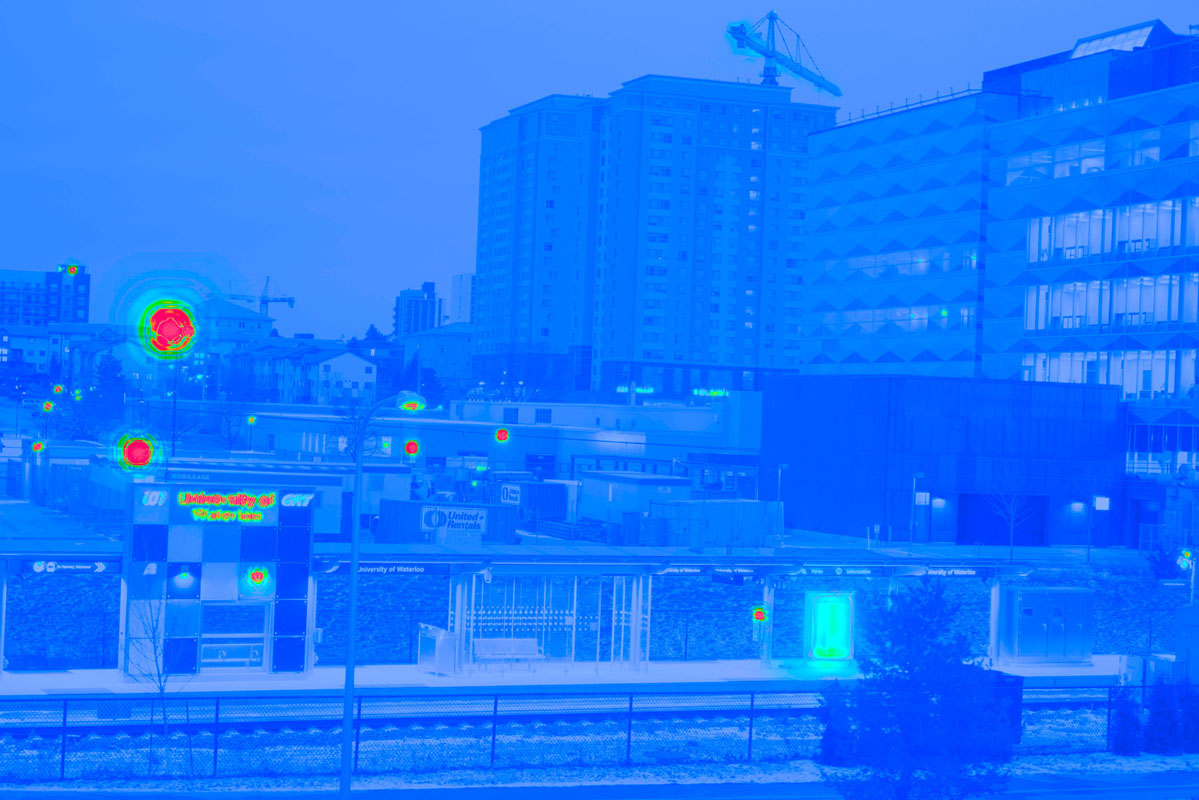}
    & \placeimage[\scale]{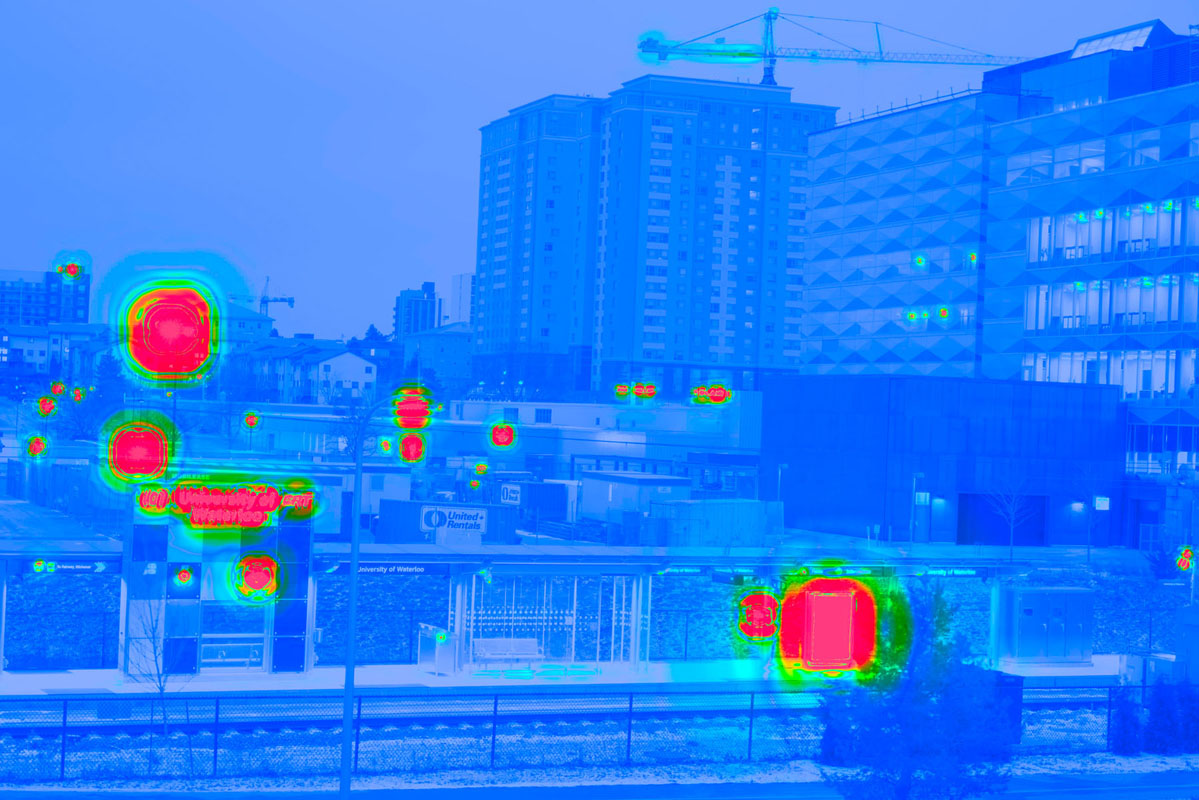}
    & \placeimage[\scale]{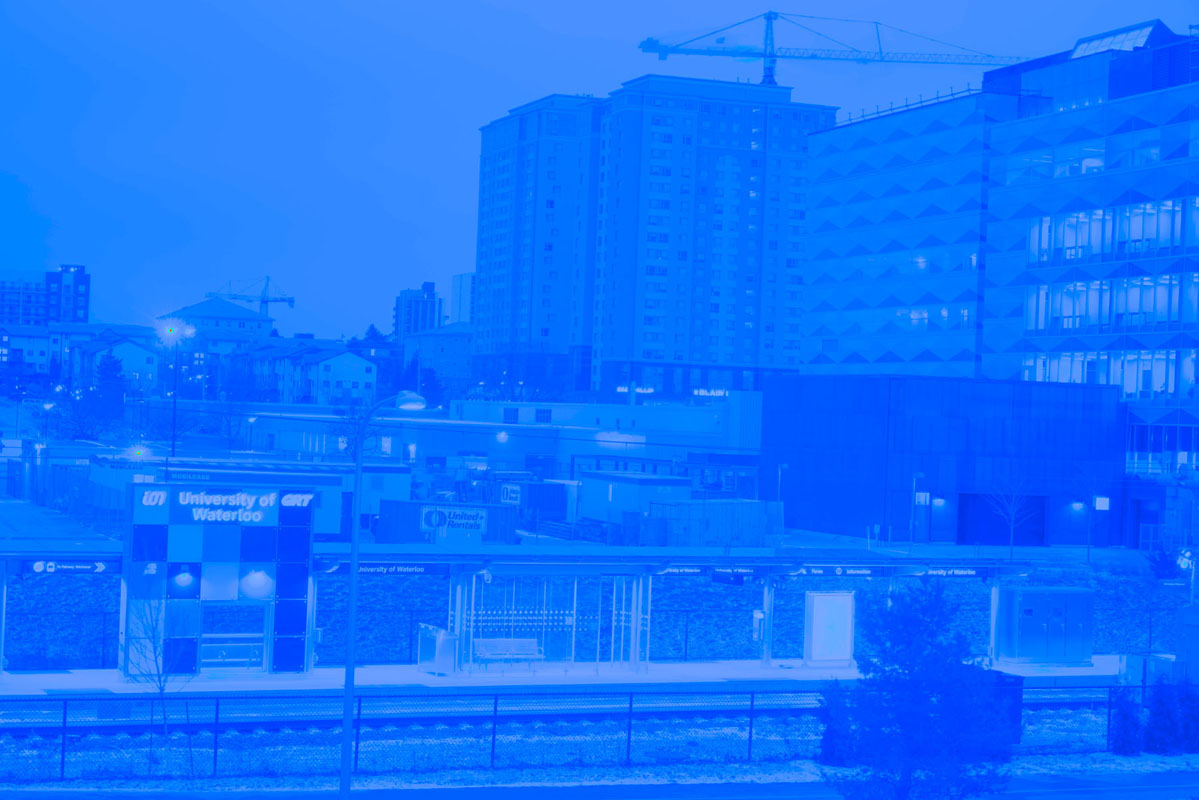} \\
    &  67.6, $2.65 \times 10^{-4}$, 4
    &  66.4, $2.77 \times 10^{-4}$, 3
    &  57.1, $3.30 \times 10^{-4}$, 2
    &  56.9, $3.31 \times 10^{-4}$, 1
    &  72.3, $2.21 \times 10^{-4}$, 6 \\
\placeimagerotate[\scale]{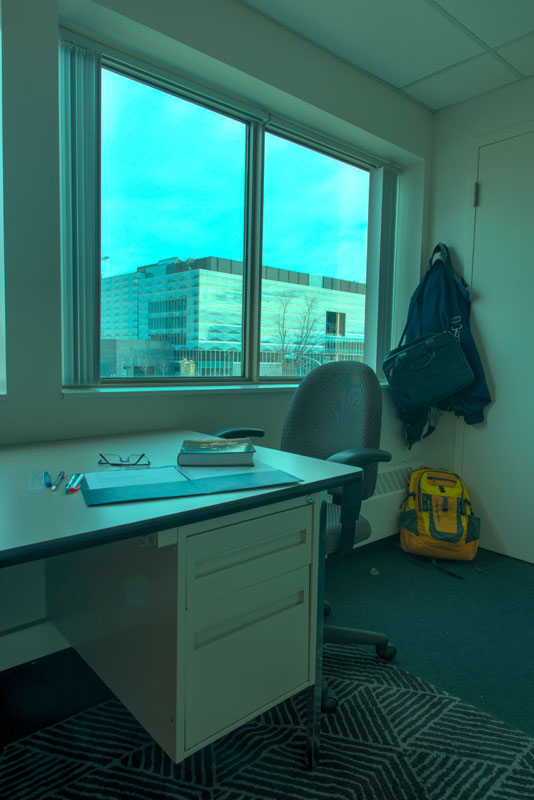}
    & \placeimagerotate[\scale]{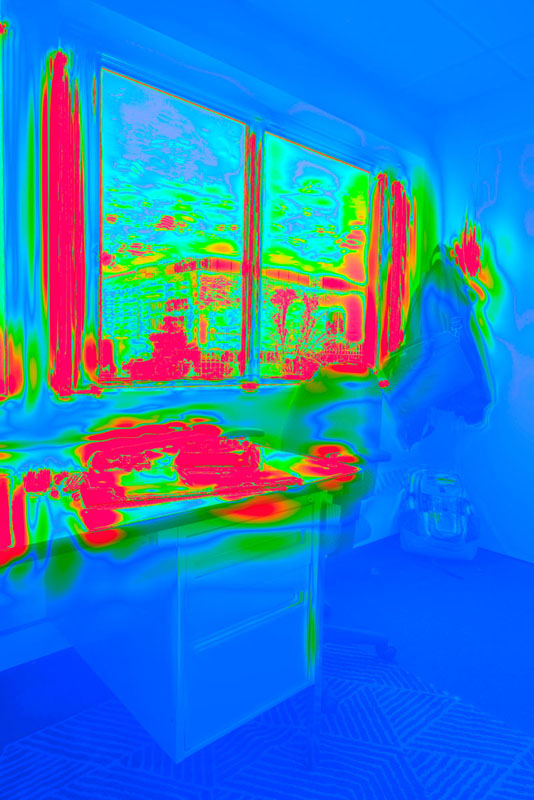}
    & \placeimagerotate[\scale]{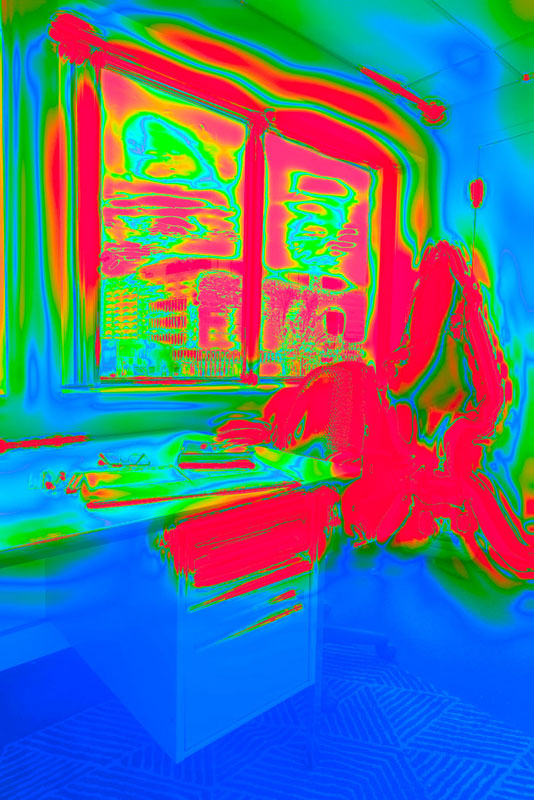}
    & \placeimagerotate[\scale]{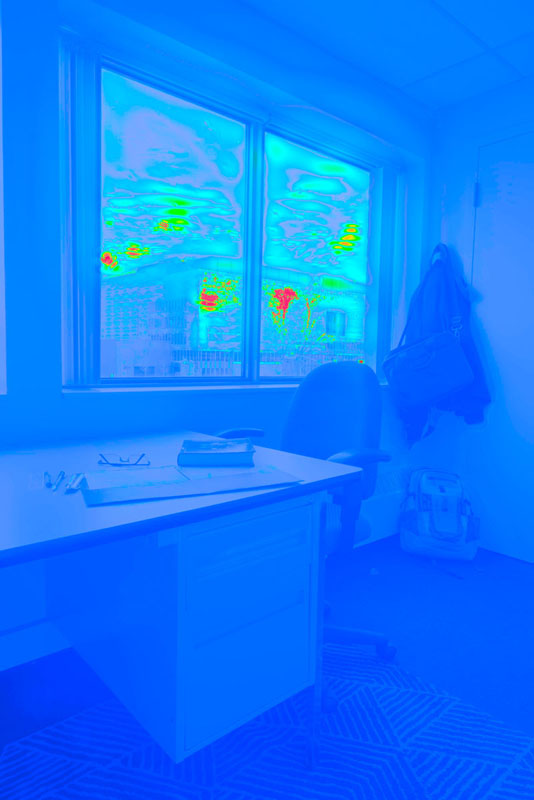}
    & \placeimagerotate[\scale]{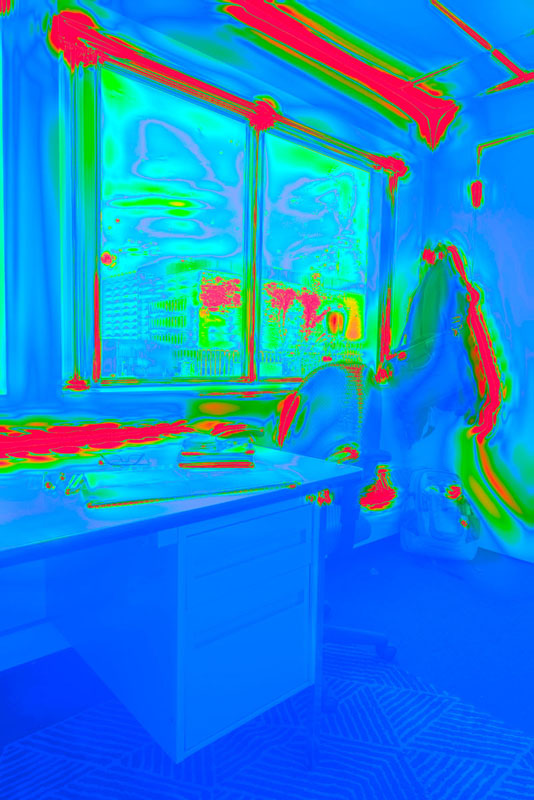}
    & \placeimagerotate[\scale]{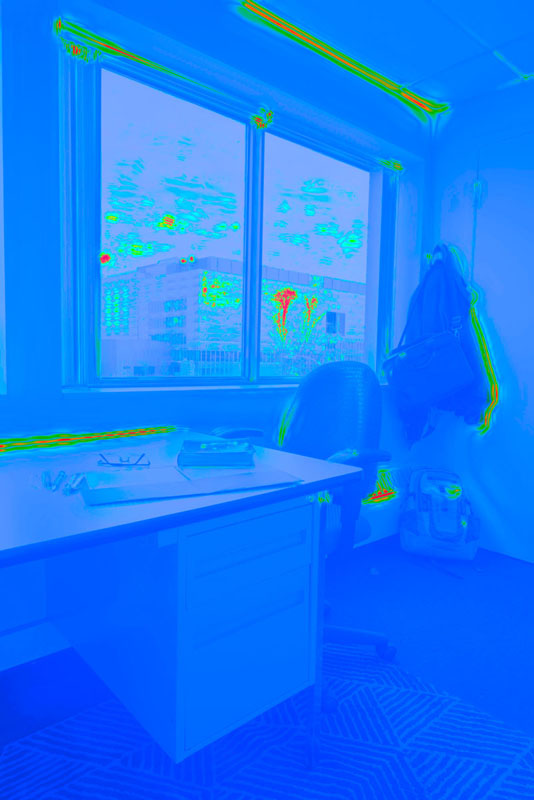} \\
    &  79.7, $ 0.905 \times 10^{-4}$, 4
    &  78.5, $ 3.00  \times 10^{-4}$, 3
    &  78.4, $ 2.76  \times 10^{-4}$, 3
    &  77.5, $ 1.98  \times 10^{-4}$, 2
    &  78.9, $ 0.108 \times 10^{-4}$, 3 \\
\placeimage[\scale]{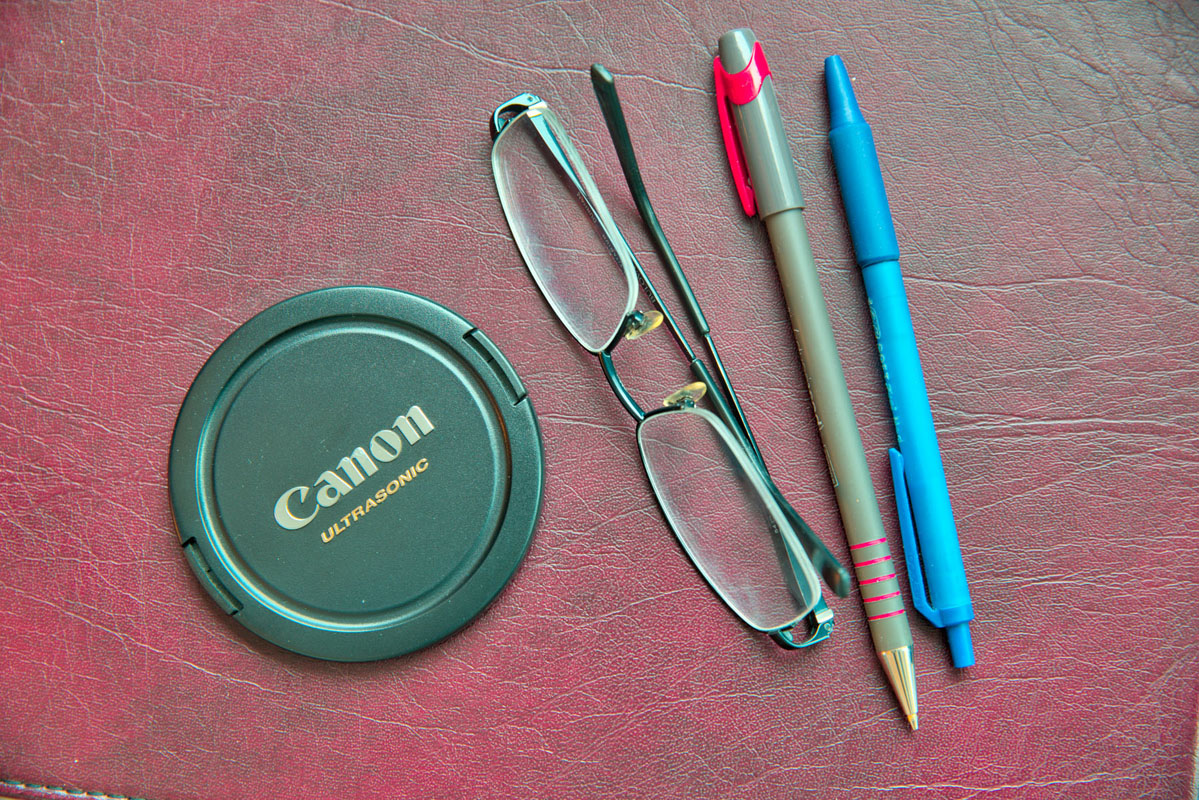}
    & \placeimage[\scale]{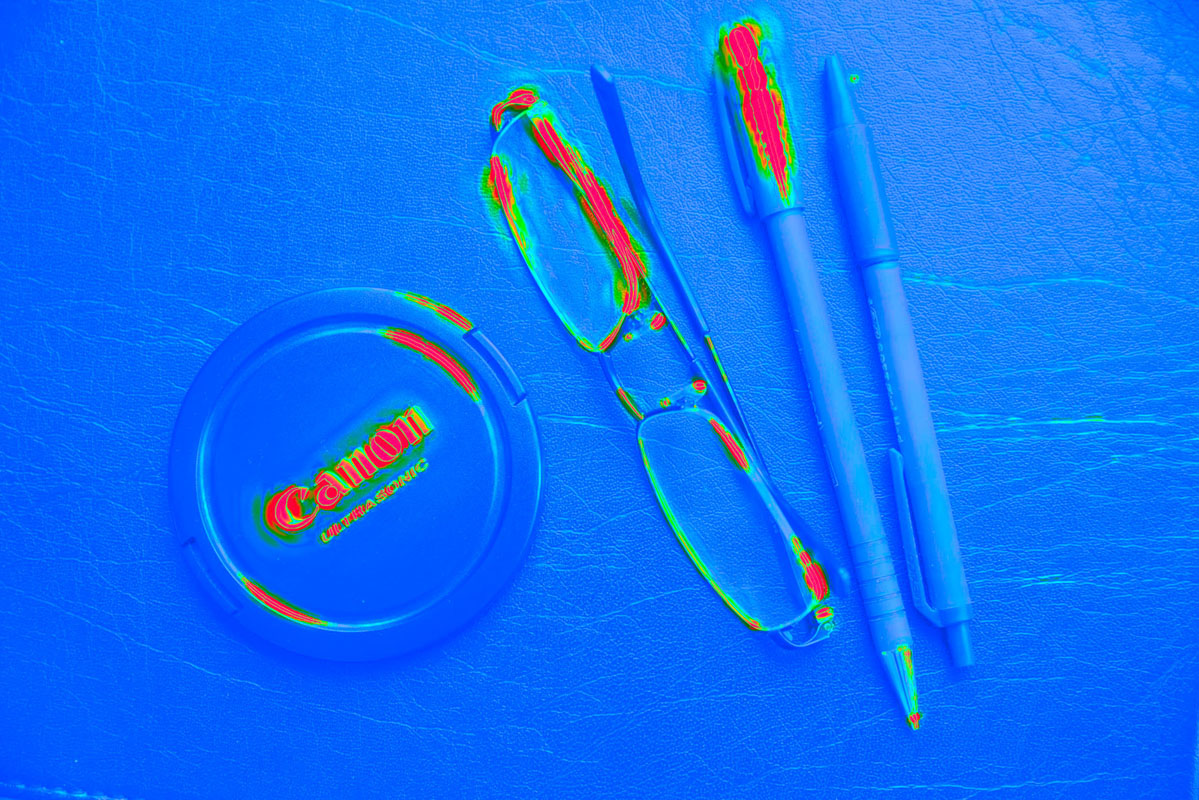}
    & \placeimage[\scale]{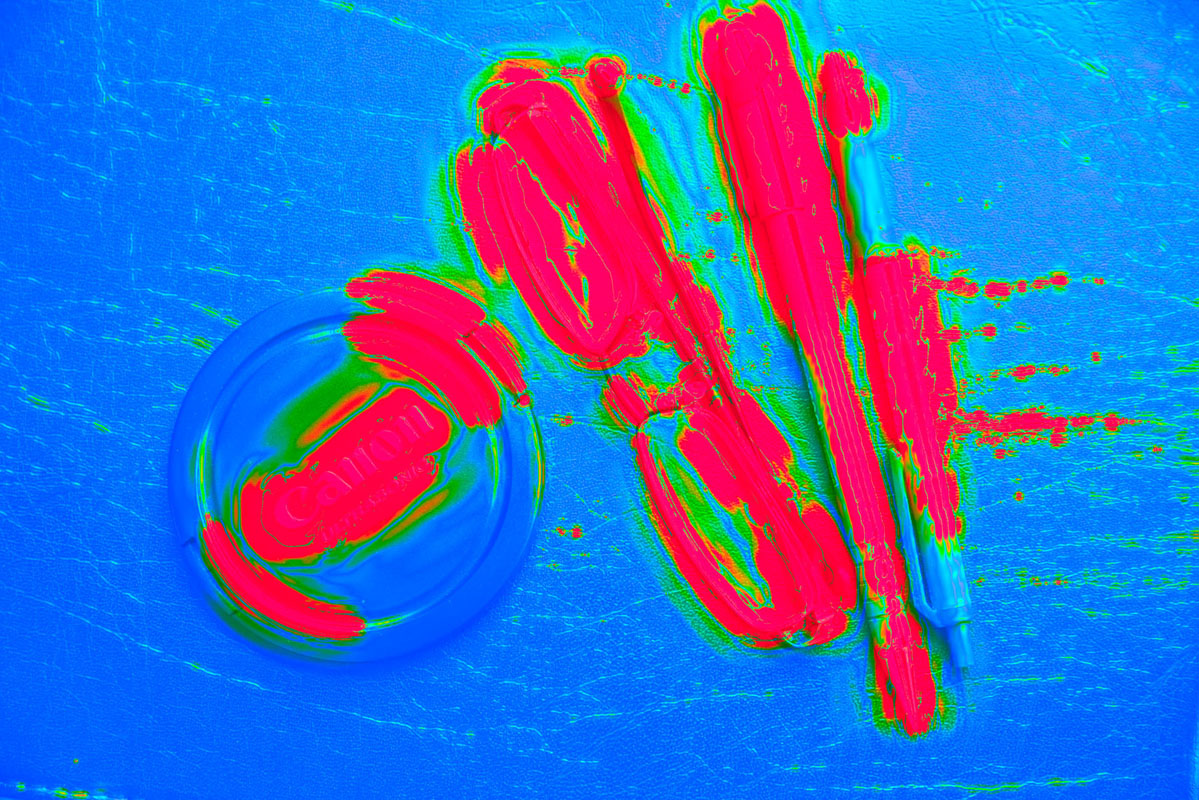}
    & \placeimage[\scale]{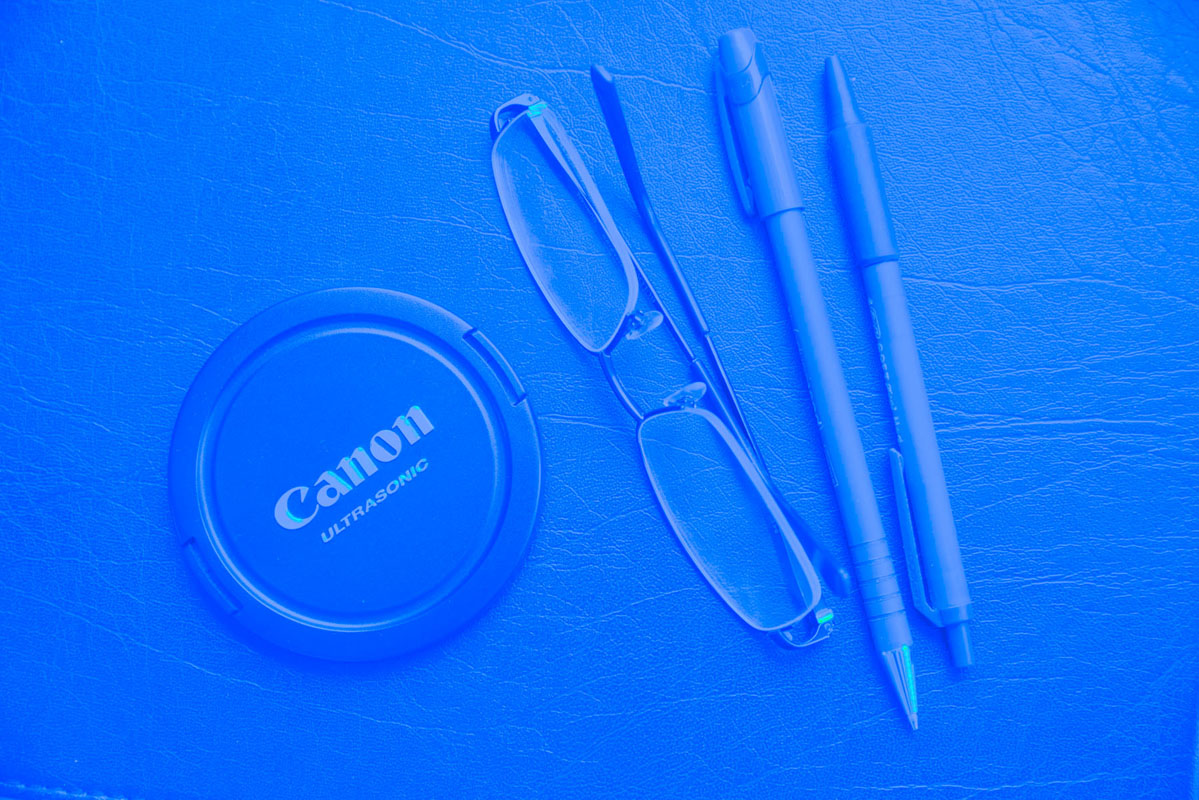}
    & \placeimage[\scale]{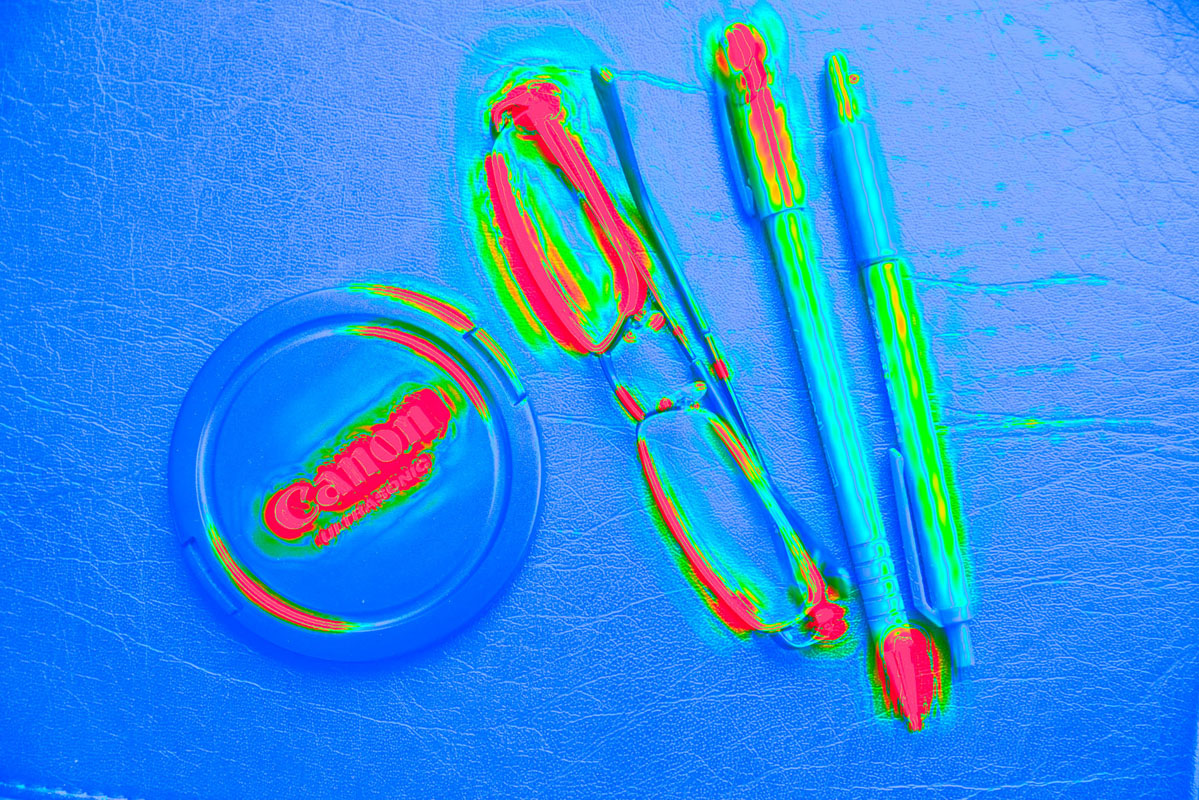}
    & \placeimage[\scale]{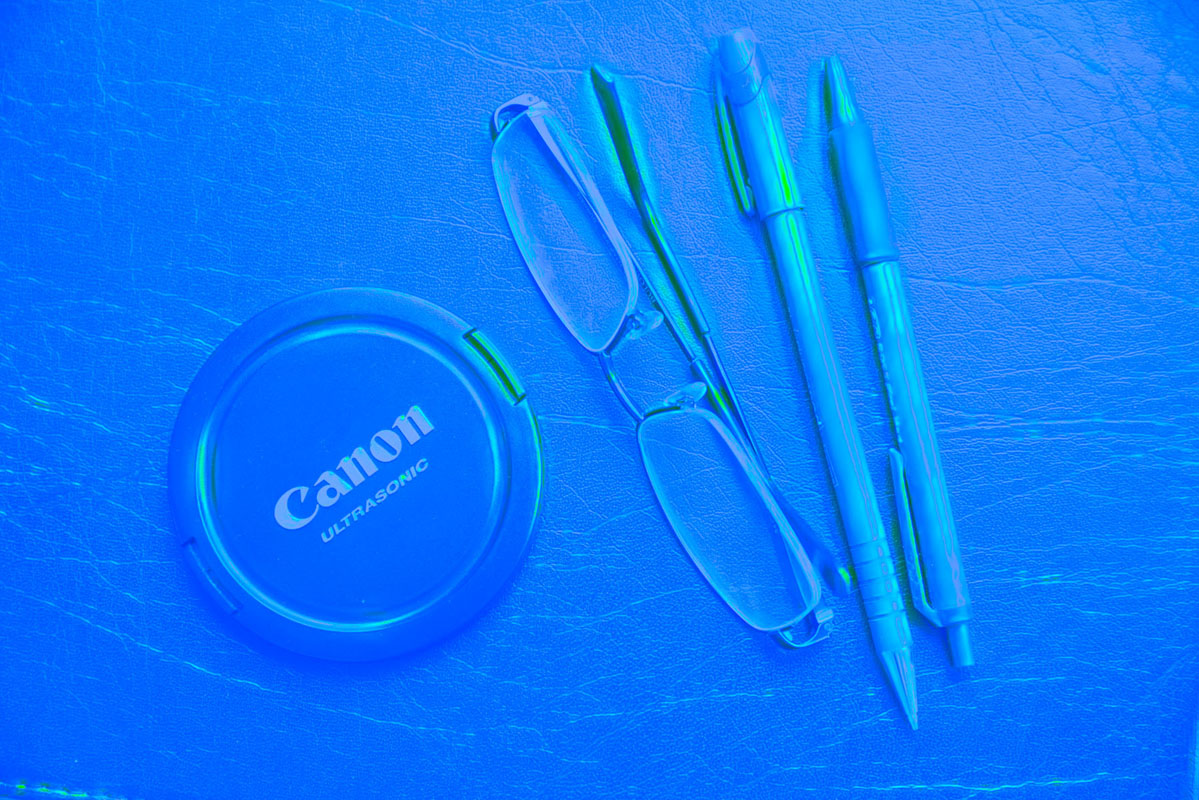} \\
    &  93.9, $0.593  \times 10^{-8}$, 3
    &  91.1, $2.04   \times 10^{-8}$, 3
    &  94.5, $0.233  \times 10^{-8}$, 4
    &  87.9, $4.96   \times 10^{-8}$, 1
    &  98.3, $0.0668 \times 10^{-8}$, 3 \\
\placeimage[\scale]{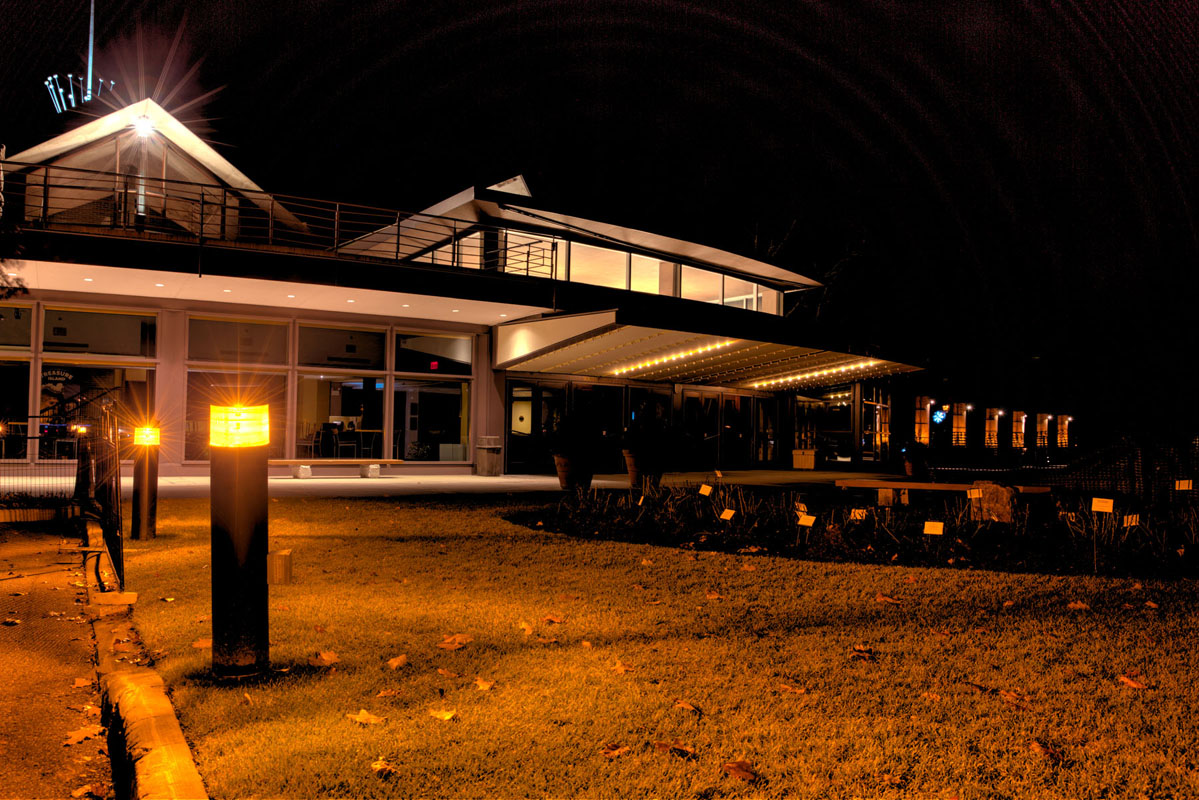}
    & \placeimage[\scale]{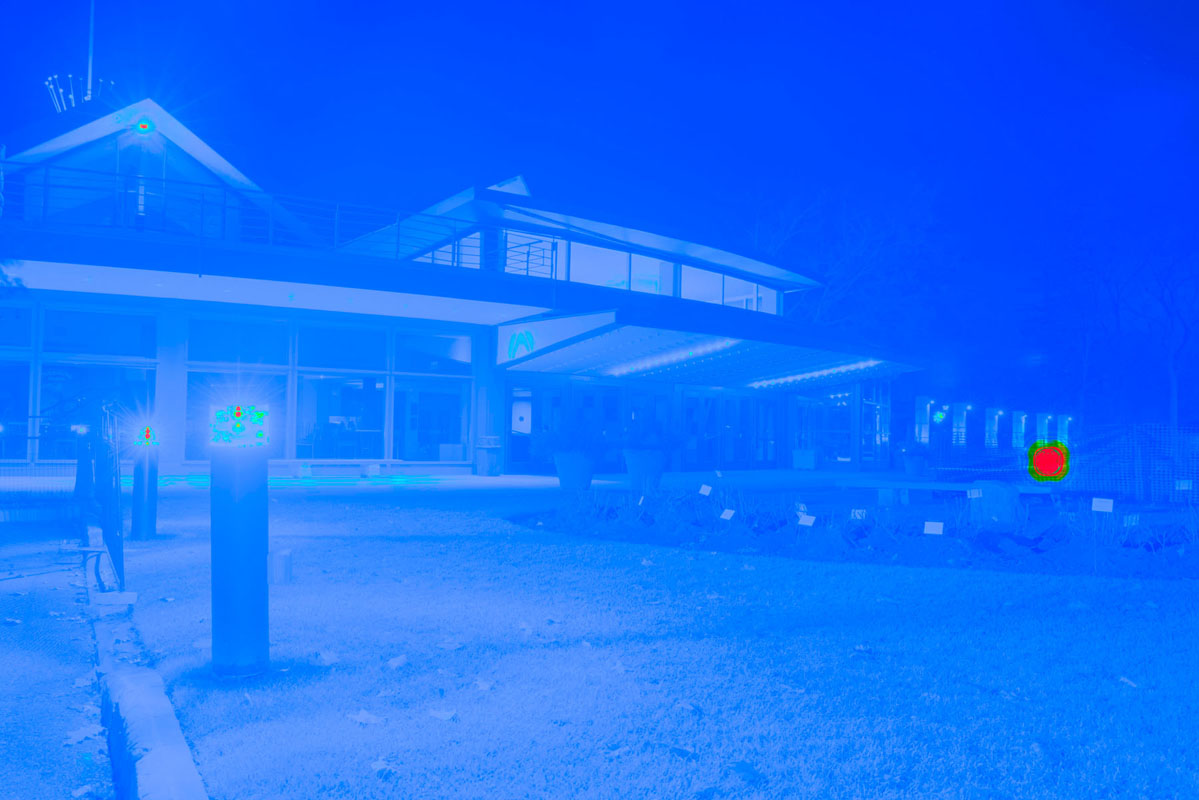}
    & \placeimage[\scale]{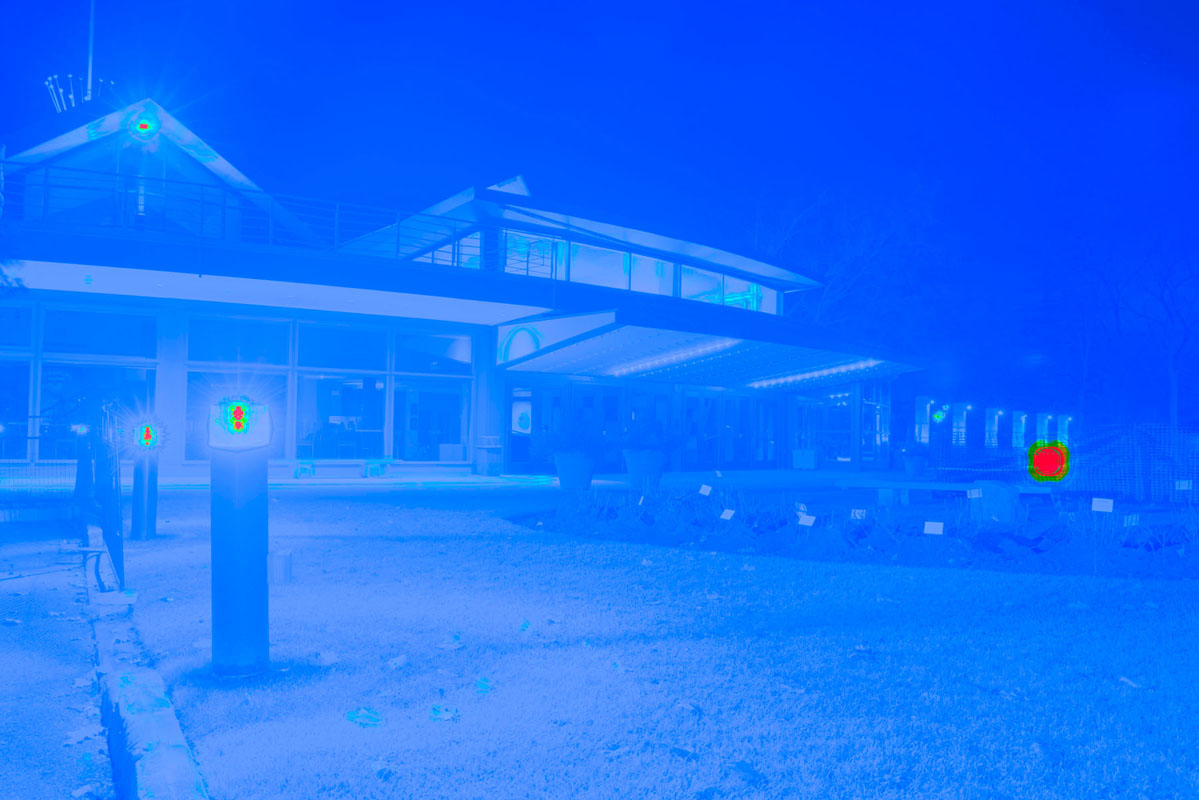}
    & \placeimage[\scale]{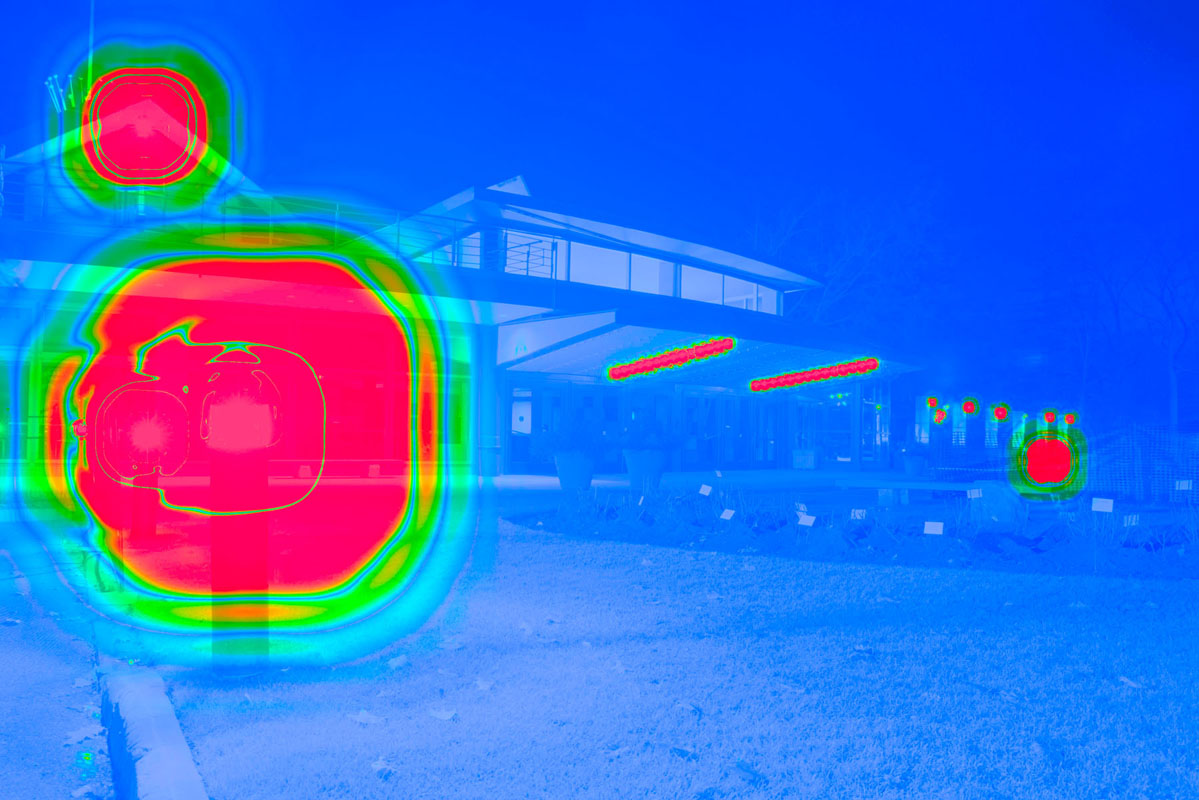}
    & \placeimage[\scale]{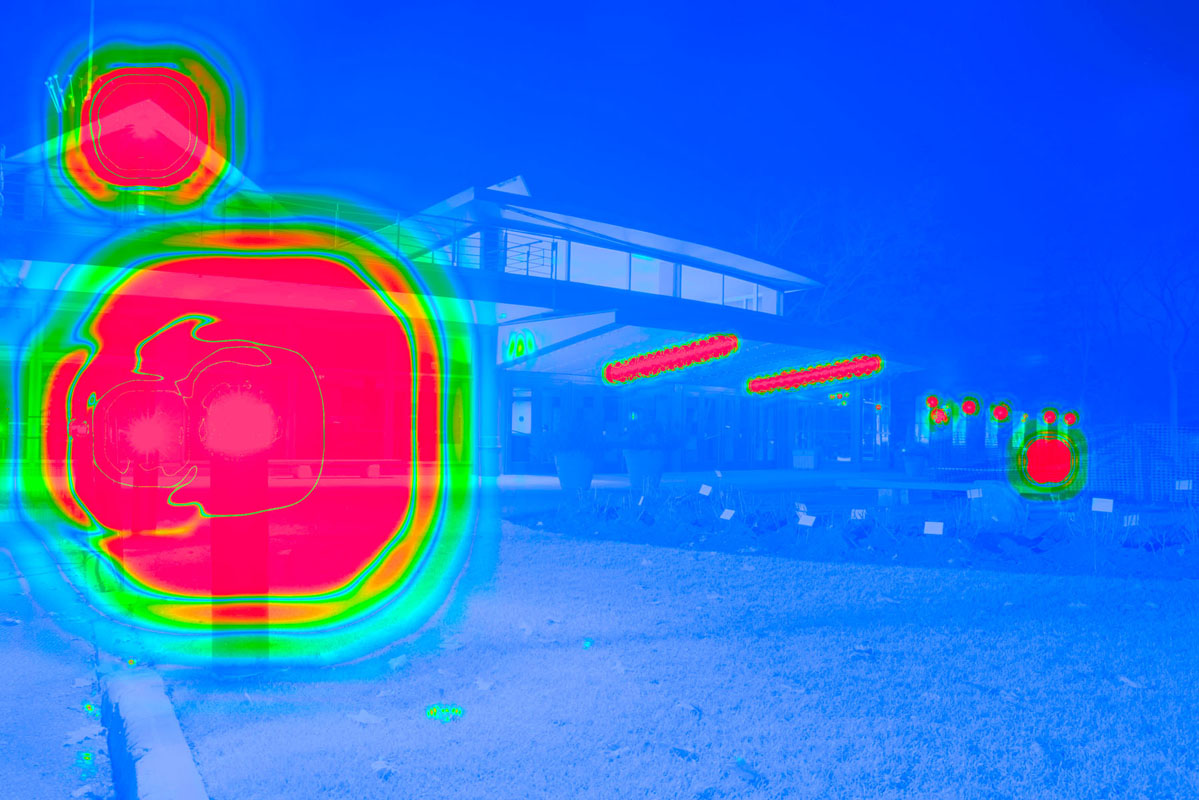}
    & \placeimage[\scale]{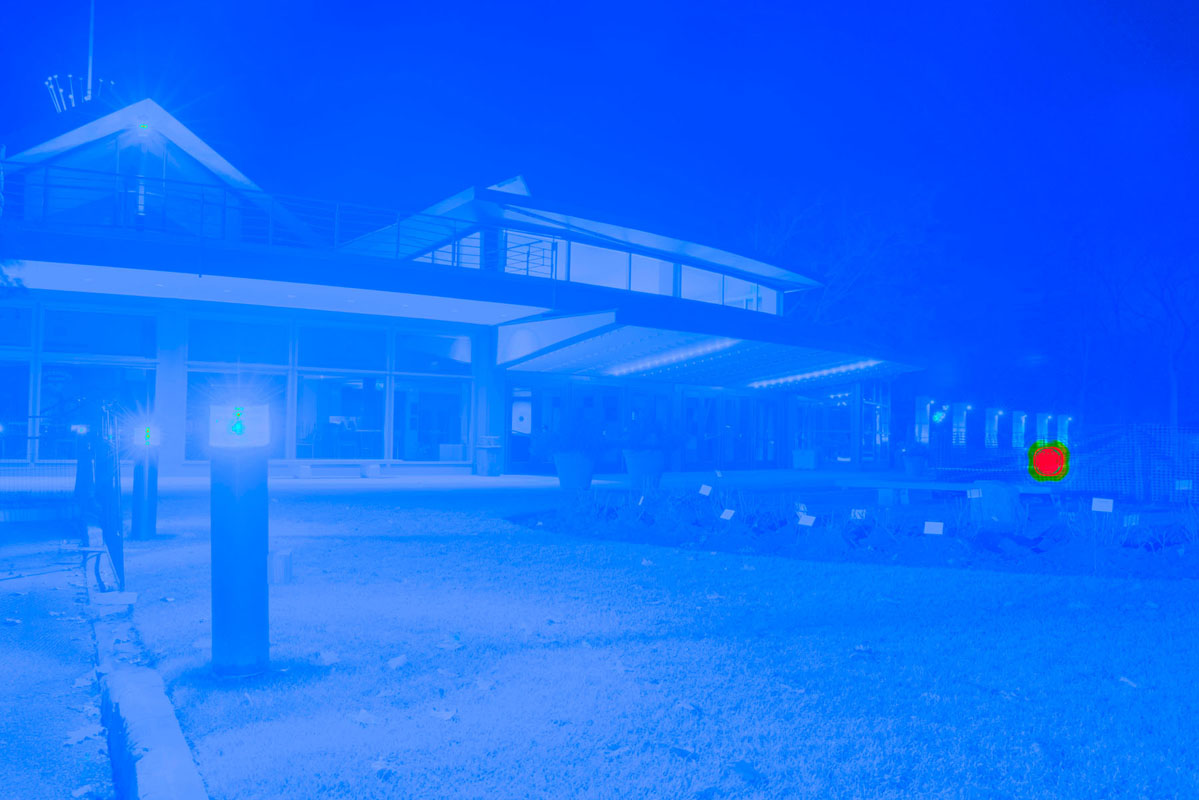} \\
    &  57.8, $5.92 \times 10^{-2}$, 5
    &  56.9, $6.42 \times 10^{-2}$, 3
    &  47.4, $6.85 \times 10^{-2}$, 3
    &  47.4, $6.85 \times 10^{-2}$, 5
    &  59.1, $2.38 \times 10^{-2}$, 7 \\
(a) & (b) & (c) & (d) & (e) & (f)
\end{tabular}
\caption{
    For various benchmark scenes,
    (a) ground truth HDR image (tonemapped),
    (b)--(f) visual representation of prediction of visually
    significant differences from ground truth \cite{Narwaria2015},
    quality correlate (/100) \cite{Narwaria2015},
    mean squared error, and
    number of images selected for various
    exposure selection methods:
    (b) Barakat et al.~\cite{Barakat2008};
    (c) Hasinoff et al.~\cite{Hasinoff2010};
    (d) Pourreza-Shahri and Kehtarnavaz \cite{Kehtarnavaz2015};
    (e) Seshadrinathan et al.~\cite{Sesh2012}; and
    (f) our proposed set covering method.
    Best viewed in color; zoom in for additional detail.
}\label{FIGURE:hdrvdp-comparison}
\end{figure}
\end{landscape}

Figure~\ref{FIGURE:hdrvdp-comparison} summarizes the results
for the five Canon benchmarks\footnote{Two of the scenes have dark regions
(second and last row).
In both of these scenes, Hasinoff et al.'s~\cite{Hasinoff2010} method 
selected exposures that included 76 multiple captures of the
longest shutter speed (32 sec.). In these cases, we restricted
the exposure set to single captures; thus the results shown for these
scenes may be somewhat worse than they would be.}.
The (non-tonemapped) images encode the probability $p$
that an average human observer would detect
a significant difference from ground truth:
blue, $p = 0.0$;
cyan, $p = 0.25$;
green, $p = 0.50$;
yellow, $p = 0.75$; and
red, $p = 1.0$.
It can be seen that on these benchmarks, with a more extensive
range of choice in each benchmark---55 images are available for selection,
each $1/3$ of a stop apart---our set covering method
consistently achieves excellent results as measured by all
three performance metrics.
On these benchmarks, the existing state-of-the-art methods
fail to fully capture the more challenging scenes, whereas our
set covering method successfully captures the scenes.

The methods can also be compared using speed as the
performance measure. The methods can be clustered into
fast (Barakat et al.~\cite{Barakat2008},
Pourreza-Shahri and Kehtarnavaz \cite{Kehtarnavaz2015}, and our
set covering), and slow (Hasinoff et al.~\cite{Hasinoff2010},
Seshadrinathan et al.~\cite{Sesh2012}). A MATLAB implementation
of our set covering method took approximately 2/3 sec.~for
pixel classification for 55 images (Step 2) and
1/3 sec.~for image set selection (Step 3). In comparison,
Hasinoff et al.'s~\cite{Hasinoff2010} method took an average of
82.4 sec. The key reason for the speed and computational
complexity differences between the fast and
slow methods is that the fast methods place a threshold on
acceptable SNR for each LDR image whereas the slow methods place
the threshold on the final HDR image. The former allows fast
greedy methods whereas the latter raises the computational
complexity significantly (either known to be NP-complete or
empirically shown to be high).


Finally, the methods can also be compared using the median
and 75\emph{th} percentile number of images selected across
all 110 benchmarks: Barakat et al.~\cite{Barakat2008}, 4 and 4
images; Hasinoff et al.~\cite{Hasinoff2010}, 3 and 3 images;
Pourreza-Shahri and Kehtarnavaz \cite{Kehtarnavaz2015},
3 and 4 images; Seshadrinathan et al.~\cite{Sesh2012}, 1
and 2 images; and our proposed set covering method, 3 and 4
images, respectively. Thus, the improvement in accuracy of
our proposed method is not at the expense of
increasing the capture time for most scenes.

%
%
\section{Conclusion}

We proposed a method for selecting the set of images to combine
in stack-based HDR imaging that is fast and offers improved accuracy. Our
technique minimizes the set of images to combine, while ensuring
that the resulting HDR image faithfully captures the scene's
irradiance. On 110 benchmark scenes, our proposed method gave
improved HDR images as measured against ground truth using
a pixel-based metric and two perception-based metrics.
As well, our experimental evaluation was the
first to extensively evaluate existing state-of-the-art methods
for image selection for stack-based HDR imaging.


\small
\bibliographystyle{IEEEbib}
\bibliography{hdr}

\begin{thebibliography}{10}

\bibitem{Mann1993}
Steve Mann and Rosalind~W. Picard,
\newblock ``On being ``undigital'' with digital cameras: Extending dynamic
  range by combining different exposed pictures,''
\newblock in {\em Proceedings of IS\&T Annual Meeting}, 1993, pp. 442--448.

\bibitem{Debevec1997}
Paul~E. Debevec and Jitendra Malik,
\newblock ``Recovering high dynamic range radiance maps from photographs,''
\newblock in {\em Proceedings of ACM SIGGRAPH}, 1997, pp. 369--378.

\bibitem{Reinhard2010}
Erik Reinhard, Greg Ward, Sumanta Pattanaik, Paul Debevec, Wolfgang Heidrich,
  and Karol Myszkowski,
\newblock {\em High Dynamic Range Imaging: Acquisition, Display, and
  Image-Based Lighting},
\newblock Morgan Kaufmann, 2nd edition, 2010.

\bibitem{Bloch2012}
Christian Bloch,
\newblock {\em The {HDRI} Handbook 2.0: High Dynamic Range Imaging for
  Photographers and CG Artists},
\newblock Rocky Nook, 2012.

\bibitem{Chen2002}
Ting Chen and Abbas El~Gamal,
\newblock ``Optimal scheduling of capture times in a multiple capture imaging
  system,''
\newblock in {\em Proceedings of Sensors and Camera Systems for Scientific,
  Industrial, and Digital Photography Applications III, SPIE Vol.~4669}, 2002.

\bibitem{Grossberg2003}
Michael~D. Grossberg and Shree~K. Nayar,
\newblock ``High dynamic range from multiple images: Which exposures to
  combine,''
\newblock in {\em Proceedings of the ICCV Workshop on Color and Photometric
  Methods in Computer Vision}, 2003, pp. 1--8.

\bibitem{Barakat2008}
Neil Barakat, A.~Nicholas Hone, and Thomas~E. Darcie,
\newblock ``Minimal-bracketing sets for high-dynamic-range image capture,''
\newblock {\em IEEE Transactions on Image Processing}, vol. 17, pp. 1864--1875,
  2008.

\bibitem{Gelfand2010}
Natasha Gelfand, Andrew Adams, Sung~Hee Park, and Kari Pulli,
\newblock ``Multi-exposure imaging on mobile devices,''
\newblock in {\em Proceedings of the ACM International Conference on
  Multimedia}, 2010, pp. 823--826.

\bibitem{Granados2010}
Miguel Granados, Boris Ajdin, Michael Wand, Christian Theobalt, Hans-Peter
  Seidel, and Hendrik~P.~A. Lensch,
\newblock ``Optimal {HDR} reconstruction with linear digital cameras,''
\newblock in {\em Proceedings of the IEEE Conference on Computer Vision and
  Pattern Recognition}, 2010, pp. 215--222.

\bibitem{Hasinoff2010}
Samuel~W. Hasinoff, Fr\'{e}do Durand, and William~T. Freeman,
\newblock ``Noise-optimal capture for high dynamic range photography,''
\newblock in {\em Proceedings of the IEEE Conference on Computer Vision and
  Pattern Recognition}, 2010, pp. 553--560.

\bibitem{Hirakawa2010}
Keigo Hirakawa and Patrick~J. Wolfe,
\newblock ``Optimal exposure control for high dynamic range imaging,''
\newblock in {\em Proceedings of the IEEE International Conference on Image
  Processing}, 2010, pp. 3137--3140.

\bibitem{Gallo2012}
Orazio Gallo, Marius Tico, Natasha Gelfand, and Kari Pulli,
\newblock ``Metering for exposure stacks,''
\newblock {\em Eurographics}, vol. 31, pp. 479--488, 2012.

\bibitem{Sesh2012}
Kalpana Seshadrinathan, Sung~Hee Park, and Oscar Nestares,
\newblock ``Noise and dynamic range optimal computational imaging,''
\newblock in {\em Proceedings of the IEEE International Conference on Image
  Processing}, 2012, pp. 2785--2788.

\bibitem{Huang2013}
Kun-Fang Huang and Jui-Chiu Chiang,
\newblock ``Intelligent exposure determination for high quality {HDR} image
  generation,''
\newblock in {\em Proceedings of the IEEE International Conference on Image
  Processing}, 2013, pp. 3201--3205.

\bibitem{Kehtarnavaz2015}
Reza Pourreza-Shahri and Nasser Kehtarnavaz,
\newblock ``Exposure bracketing via automatic exposure selection,''
\newblock in {\em Proceedings of the IEEE International Conference on Image
  Processing}, 2015, pp. 320--323.

\bibitem{Chakrabarti2009}
Ayan Chakrabarti, Daniel Scharstein, and Todd Zickler,
\newblock ``An empirical camera model for internet color vision,''
\newblock in {\em Proceedings of the British Machine Vision Conference}, 2009.

\vfill\pagebreak
\bibitem{Szeliski2010}
Richard Szeliski,
\newblock {\em Computer Vision: Algorithms and Applications},
\newblock Springer, 2010.

\bibitem{Healey1994}
Glenn~E. Healey and Raghava Kondepudy,
\newblock ``Radiometric {CCD} camera calibration and noise estimation,''
\newblock {\em IEEE Transactions on Pattern Analysis and Machine Intelligence},
  vol. 16, pp. 267--276, 1994.

\bibitem{Mitsunaga1999}
Tomoo Mitsunaga and Shree~K. Nayar,
\newblock ``Radiometric self calibration,''
\newblock in {\em Proceedings of the IEEE Conference on Computer Vision and
  Pattern Recognition}, 2005, pp. 374--380.

\bibitem{Kim2012}
Seon~Joo Kim, Hai~Ting Lin, Zheng Lu, Sabine S\"{u}sstrunk, Stephen Lin, and
  Michael~S. Brown,
\newblock ``A new in-camera imaging model for color computer vision and its
  application,''
\newblock {\em IEEE Transactions on Pattern Analysis and Machine Intelligence},
  vol. 34, pp. 2289--2302, 2012.

\bibitem{Liu2008}
Ce~Liu, Richard Szeliski, Sing~Bing Kang, C.~Lawrence Zitnick, and William~T.
  Freeman,
\newblock ``Automatic estimation and removal of noise from a single image,''
\newblock {\em IEEE Transactions on Pattern Analysis and Machine Intelligence},
  vol. 30, pp. 299--314, 2008.

\bibitem{VaqueroGTPT2011}
Daniel Vaquero, Natasha Gelfand, Marius Tico, Kari Pulli, and Matthew Turk,
\newblock ``Generalized autofocus,''
\newblock in {\em Proceedings of the IEEE Workshop on Applications of Computer
  Vision}, 2011.

\bibitem{GareyJ79}
Michael~R. Garey and David~S. Johnson,
\newblock {\em Computers and Intractability: A Guide to the Theory of
  NP-Completeness},
\newblock W. H. Freeman, 1979.

\bibitem{Nemhauser1988}
George~L. Nemhauser and Laurence~A. Wolsey,
\newblock {\em Integer and Combinatorial Optimization},
\newblock Wiley, 1988.

\bibitem{Schobel2004}
Anita Sch\"{o}bel,
\newblock ``Set covering problems with consecutive ones property,''
\newblock Technical report, Universit\"{a}t Kaiserslautern, 2004.

\bibitem{Kirk2006}
Kristian Kirk and Hans~J{\o}rgen Andersen,
\newblock ``Noise characterization of weighting schemes for combination of
  multiple exposures,''
\newblock in {\em Proceedings of the British Machine Vision Conference}, 2006.

\bibitem{Robertson2003}
Mark~A. Robertson, Sean Borman, and Robert~L. Stevenson,
\newblock ``Estimation-theoretic approach to dynamic range enhancement using
  multiple exposures,''
\newblock {\em Journal of Electronic Imaging}, vol. 12, pp. 219--228, 2003.

\bibitem{Mecke2004}
Steffen Mecke and Dorothea Wagner,
\newblock ``Solving geometric covering problems by data reduction,''
\newblock in {\em Proc.~of European Symposia on Algorithms}, 2004, pp.
  760--771.

\bibitem{Fairchild2007}
Mark~D. Fairchild,
\newblock ``The {HDR} photographic survey,''
\newblock in {\em Color and Imaging Conference}, 2007, pp. 233--238.

\bibitem{Mantiuk2011}
Rafat Mantiuk, Kil~Joong Kim, Allan~G. Rempel, and Wolfgang Heidrich,
\newblock ``{HDR-VDP-2}: A calibrated visual metric for visibility and quality
  predictions in all luminance conditions,''
\newblock {\em ACM Transactions on Graphics}, vol. 30, no. 4, pp. 40:1--40:14,
  2011.

\bibitem{Narwaria2015}
Manish Narwaria, Rafal~K. Mantiuk, Mattheiu~Perreira Da~Silva, and Patrick
  Le~Callet,
\newblock ``{HDR-VDP-2.2}: A calibrated method for objective quality prediction
  of high dynamic range and standard images,''
\newblock {\em Journal of Electronic Imaging}, vol. 24, 2015.

\end{thebibliography}


\begin{biography}
Peter van Beek received a BSc from the University of British Columbia
(1984) and an MMath and PhD in Computer Science from the University of
Waterloo (1986, 1990). He has been a faculty member in Computer Science
at the University of Waterloo since 2000. In 2008, he was named a
Fellow of the Association for Artificial Intelligence. His research
interests span the field of AI with a focus on optimization,
constraint programming, and applied machine learning.
\end{biography}

\end{document}